\documentclass[sigconf]{aamas}

\usepackage{balance} 

\usepackage{algorithm}
\usepackage{algorithmic}
\usepackage{url}
\usepackage{wrapfig}
\usepackage[utf8]{inputenc} 
\usepackage{amsfonts}       
\usepackage{nicefrac}       
\usepackage{xcolor}         
\usepackage[most]{tcolorbox}
\usepackage{lipsum}
\usepackage{listings}
\definecolor{codegreen}{rgb}{0,0.6,0}
\definecolor{codegray}{rgb}{0.5,0.5,0.5}
\definecolor{codepurple}{rgb}{0.58,0,0.82}
\definecolor{backcolour}{rgb}{0.95,0.95,0.92}
\usepackage{textcomp} 
\usepackage[T1]{fontenc}
\lstdefinestyle{mystyle}{
    language=Python,
    backgroundcolor=\color{black!5},   
    commentstyle=\color{green!40!black},
    keywordstyle=\color{blue},
    stringstyle=\color{purple},
    numberstyle=\tiny\color{gray},
    basicstyle=\footnotesize\ttfamily, 
    breakatwhitespace=true,          
    breaklines=true,                 
    captionpos=b,                    
    frame=single,                    
    keepspaces=true,                 
    numbers=left,                    
    numbersep=5pt,                   
    showstringspaces=false,          
    tabsize=4,                       
    upquote=true,                    
    postbreak=\mbox{\textcolor{red}{$\hookrightarrow$}\space} 
}

\setcopyright{ifaamas}
\acmConference[AAMAS '26]{Proc.\@ of the 25th International Conference
on Autonomous Agents and Multiagent Systems (AAMAS 2026)}{May 25 -- 29, 2026}
{Paphos, Cyprus}{C.~Amato, L.~Dennis, V.~Mascardi, J.~Thangarajah (eds.)}
\copyrightyear{2026}
\acmYear{2026}
\acmDOI{}
\acmPrice{}
\acmISBN{}

\acmSubmissionID{7}

\title[COMPASS]{Closed-Loop Vision-Language Planning for Multi-Agent Coordination}

\author{Zhiyuan Li}
\affiliation{
  \institution{Department of Electrical Engineering and Automation, Aalto University}
  \country{Finland}}
\email{zhiyuan.li@aalto.fi}

\author{Wenshuai Zhao}
\affiliation{
  \institution{Department of Computer Science, Aalto University}
  \country{Finland}}
\email{wenshuai.zhao@aalto.fi}

\author{Joni Pajarinen}
\affiliation{
  \institution{Department of Electrical Engineering and Automation, Aalto University}
  \country{Finland}}
\email{joni.pajarinen@aalto.fi}

\begin{abstract}
Cooperative multi-agent reinforcement learning (MARL) struggles with sample efficiency, interpretability, and generalization. While Large Language Models (LLMs) offer powerful planning capabilities, their application has been hampered by a reliance on text-only inputs and a failure to handle the non-Markovian, partially observable nature of multi-agent tasks. We introduce COMPASS, a multi-agent framework that overcomes these limitations by integrating Vision-Language Models (VLMs) for decentralized, closed-loop decision-making. COMPASS dynamically generates and refines interpretable, code-based strategies stored in a skill library that is bootstrapped from expert demonstrations. To ensure robust coordination, it propagates entity information through a structured multi-hop communication protocol, allowing teams to build a coherent understanding from partial observations. Evaluated on the challenging SMACv2 benchmark, COMPASS significantly outperforms state-of-the-art MARL baselines. Notably, in the symmetric Protoss 5v5 task, COMPASS achieved a 57\% win rate, a 30 percentage point advantage over QMIX (27\%). Project page can be found at https://stellar-entremet-1720bb.netlify.app/.
\end{abstract}


\newcommand{\BibTeX}{\rm B\kern-.05em{\sc i\kern-.025em b}\kern-.08em\TeX}

\begin{document}


\pagestyle{fancy}
\fancyhead{}


\maketitle 


\section{Introduction}
\label{introduction}
A major long-term goal for the field of cooperative multi-agent systems (MAS), e.g. multi-robot control \cite{gao2024coohoilearningcooperativehumanobject,feng2024learningmultiagentlocomanipulationlonghorizon}, power management \cite{monroc2024wfcrl}, and multi-agent games \cite{10.5555/3306127.3332052,Kurach_Raichuk_Stańczyk_Zając_Bachem_Espeholt_Riquelme_Vincent_Michalski_Bousquet_Gelly_2020}, is to build a protocol of collaboration among the agents. Multi-agent reinforcement learning (MARL), proven as an advanced paradigm of distributed artificial intelligence (AI), holds promise for discovering collective behavior from interactions. One line of works follows centralized training decentralized execution (CTDE) paradigm \cite{JMLR:v21:20-081,NEURIPS2022_9c1535a0,liu2024maximum,10.5555/3295222.3295385,zhu2025madiffofflinemultiagentlearning,Li_Zhao_Wu_Pajarinen_2024}. CTDE assumes a central controller that exploits global information, while the individual policies are designed to allow for decentralized execution. However, in many real-world scenarios, the central controller becomes unfeasible due to the communication overhead that exponentially scales with the number of agents, thereby compromising the scalability of MAS. In contrast, the decentralized training decentralized execution paradigm (DTDE) discards this assumption and is scalable to large-scale systems \cite{su2024a,pmlr-v80-zhang18n,ma2024efficient}. However, DTDE requires complicated learning and planning under uncertainty, as partial observability magnifies the discrepancy between each agent’s local observation and global information. Although much progress has been made, MARL suffers from compromised sample efficiency, interpretability, and transferability.

The emergence of Large Language Models (LLMs) has revitalized this field. LLM-based multi-agents have been proposed to leverage their remarkable capacity to perform task-oriented collective behaviors \cite{10610855,zhang2024buildingcooperativeembodiedagents,gong-etal-2024-mindagent,10.1609/aaai.v38i16.29710,nayak2025llamarlonghorizonplanningmultiagent}. The LLMs are used for high-level planning to generate centralized \cite{gong-etal-2024-mindagent,nayak2025llamarlonghorizonplanningmultiagent,deng2024newapproachsolvingsmac} or decentralized plans \cite{10610855,zhang2024buildingcooperativeembodiedagents}, often adopting a hierarchical decision-making structure in conjunction with a pre-defined low-level controller. While these methods have succeeded in a set of multi-agent problems including Overcooked-AI \cite{10.5555/3454287.3454752}, SMAC \cite{10.5555/3306127.3332052}, and VirtualHome \cite{Puig_2018_CVPR}, heavy reliance on text-based observation prevents them from learning from multi-modal information. Moreover, they ignore the non-Markovian nature of MAS, where learning and planning necessitate a decentralized closed-loop solution.

\begin{figure*}[t]
\centering
\includegraphics[width=\linewidth]{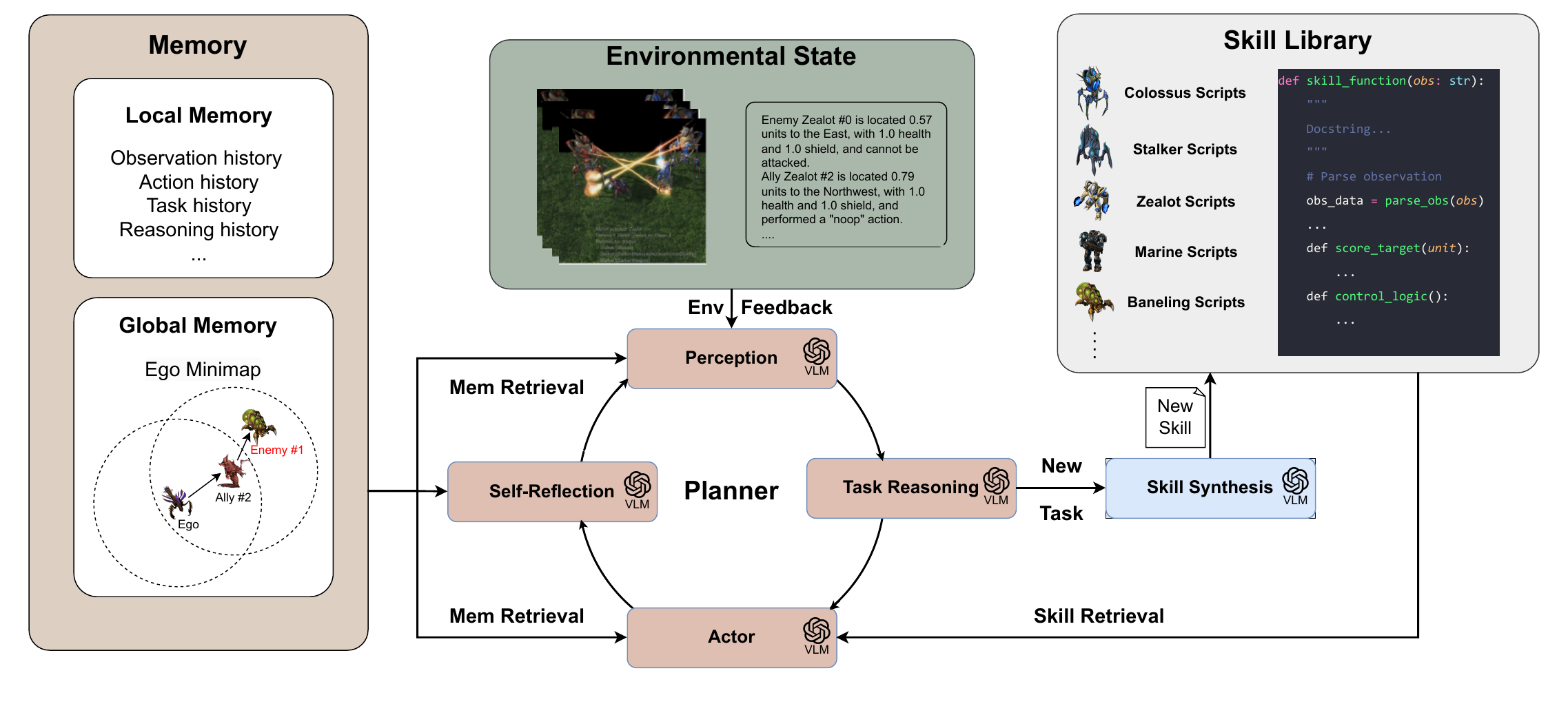}
\caption{Overview of the COMPASS architecture, a novel framework that advances cooperative multi-agent decision-making through three synergistic components: (1) A VLM-based closed-loop planner that enables decentralized control by continuously processing multi-modal feedback and adapting strategies, addressing the non-Markovian challenge of multi-agent systems; (2) A dynamic skill synthesis mechanism that combines demonstration bootstrapping with incremental skill generation, improving sample efficiency and interpretability; and (3) A structured communication protocol that facilitates efficient information sharing through entity-based multi-hop propagation, enhancing cooperative perception under partial observability.}
\Description{Overview of the COMPASS architecture, a novel framework that advances cooperative multi-agent decision-making through three synergistic components: (1) A VLM-based closed-loop planner that enables decentralized control by continuously processing multi-modal feedback and adapting strategies, addressing the non-Markovian challenge of multi-agent systems; (2) A dynamic skill synthesis mechanism that combines demonstration bootstrapping with incremental skill generation, improving sample efficiency and interpretability; and (3) A structured communication protocol that facilitates efficient information sharing through entity-based multi-hop propagation, enhancing cooperative perception under partial observability.}
\label{fig:compass}
\end{figure*}

In this paper, we mitigate the existing research limitations and advance general decision-making for cooperative multi-agent systems. At a high level, COMPASS combines a vision language models (VLMs)-based planner with a dynamic skill library for storing and retrieving complex behaviors, along with a structured communication protocol. A diagram of COMPASS is provided in Figure \ref{fig:compass}. Inspired by Cradle \cite{tan2024cradleempoweringfoundationagents}, the VLM-based planner perceives the visual and textual observation and suggests the most suitable executable code from the skill library. We adopt the code-as-policy paradigm \cite{wang2024executablecodeactionselicit} instead of task-specific primitive actions, as it constrains generalizability and fails to fully leverage foundation models' extensive world knowledge and sophisticated reasoning capabilities. 

Traditional open-loop methods struggle to produce effective plans that adapt to dynamics in stochastic, partially observable environments. To address this challenge, the VLM-based planer attempts to solve challenging and ambiguous final tasks, such as "Defeat all enemy units in the StarCraft multi-agent combat scenario while coordinating with allied units", by progressively proposing a sequence of clear, manageable sub-tasks while incorporating environmental feedback and task progress. COMPASS generates Python scripts through LLMs as semantic-level skills to accomplish sub-tasks, incrementally building a skill library throughout the task progress. Each skill is indexed through its documentation embeddings, enabling retrieval based on task-skill relevance. However, developing the skill library from scratch requires extensive exploration to discover viable strategies. In contrast, we pre-collect demonstration videos and introduce a "warm start" by initializing the skill library with strategies derived from the expert-level dataset.

Moreover, building autonomous agents to cooperate in completing tasks under partial observation requires an efficient communication protocol. However, naive communication leads to the risks of hallucination caused by meaningless chatter between agents \cite{10.5555/3666122.3668386}. Inspired by entity-based MARL \cite{pmlr-v139-iqbal21a,pmlr-v202-ding23d}, we present a structured communication protocol to formulate the communication among agents along with a global memory that allows all agents to retrieve. The protocol incorporates a multi-hop propagation mechanism, enabling agents to infer information about entities beyond their field of view through information shared by teammates. Similar to previous approaches, each agent maintains a local memory to preserve current and historical experiences.

Empirically, COMPASS demonstrates effective adaptation and skill synthesis in cooperative multi-agent scenarios. Through its dynamic skill library, it creates reusable and interpretable code-based behaviors that evolve during task execution. We evaluate COMPASS systematically in the improved StarCraft Multi-Agent Challenge (SMACv2) using both open-source (Qwen2-VL-72B \cite{wang2024qwen2vlenhancingvisionlanguagemodels}) and closed-source (GPT-4o-mini\footnote{https://platform.openai.com/docs/models\#gpt-4o-mini}, Claude-3-Haiku\footnote{https://www.anthropic.com/news/claude-3-haiku}) VLMs. COMPASS achieves strong results in Protoss scenarios with a 57\% win rate, substantially outperforming state-of-the-art MARL algorithms, including QMIX \cite{JMLR:v21:20-081}, MAPPO \cite{NEURIPS2022_9c1535a0}, HAPPO \cite{kuba2022trustregionpolicyoptimisation}, and HASAC \cite{liu2024maximum}. COMPASS maintains moderate performance in Terran scenarios and handles asymmetric settings effectively, though showing limited success in Zerg task. We evaluate the contribution of individual COMPASS components to its overall performance. We further demonstrate COMPASS's ability to bootstrap effective strategies from expert demonstrations.

\section{Related Work}
\label{related work}
\subsection{Agents in the StarCraft Multi-Agent Challenge}
SMAC \cite{10.5555/3306127.3332052}, a predominant cooperative MARL benchmark based on the StarCraft II real-time strategy game \cite{vinyals2017starcraftiinewchallenge}, focuses on decentralized micromanagement scenarios where each unit operates under decentralized execution with partial observability to defeat enemy units controlled by Starcraft II’s built-in AI opponent. Previous research in SMAC resort to MARL which can be divided into two categories: 1) Online MARL: One line of representative research is value decomposition (VD) \cite{JMLR:v21:20-081,wang2021qplex,pmlr-v97-son19a}, which decomposes the centralized action-value function into individual utility functions. On the other hand, multi-agent policy gradient (MAPG) methods \cite{NEURIPS2022_9c1535a0,kuba2022trustregionpolicyoptimisation,liu2024maximum,Li_Liu_Zhang_Wei_Niu_Yang_Liu_Ouyang_2023,NEURIPS2022_69413f87,hu2024learning,na2024efficient} extend single-agent policy gradient algorithm to multi-agent with coordination modeling. Researches such as MAPPO \cite{NEURIPS2022_9c1535a0}, HAPPO \cite{kuba2022trustregionpolicyoptimisation}, and HASAC \cite{liu2024maximum} combine trust region and maximum entropy with MARL in a non-trivial way respectively. To encourage coordination, communication methods \cite{hu2024learning,lo2024learning}, sequential modeling methods \cite{Li_Liu_Zhang_Wei_Niu_Yang_Liu_Ouyang_2023,NEURIPS2022_69413f87}, and cooperative exploration methods \cite{10.5555/3454287.3454971,na2024efficient} have been proposed. 2) Offline MARL: Recent efforts such as MADT \cite{meng2023offline}, ODIS \cite{zhang2022discovering}, and MADiff \cite{zhu2025madiffofflinemultiagentlearning} leverage data-driven training via pre-collected offline datasets to enhance policy training efficiency. However, the near-optimal performance of these existing approaches on SMAC highlights the benchmark's limited stochasticity and partial observability. To address these limitations, SMACv2 \cite{10.5555/3666122.3667756} introduces more complexity to necessitate decentralized closed-loop control policies. There have been some recent attempts \cite{Li_Zhao_Wu_Pajarinen_2024,mcclellan2024boostingsampleefficiencygeneralization,10.5555/3545946.3598961,li2024agentmixermultiagentcorrelatedpolicy} to evaluate MARL algorithms on SMACv2, and the results confirm the complexity. However, current learning-based multi-agent methods are computationally inefficient and non-interpretable. In the quest to find methods that are sample-efficient and interpretable, LLM-SMAC \cite{deng2024newapproachsolvingsmac} leverage LLMs to generate centralized decision tree code under global information in an open-loop framework. Unlike prior works, COMPASS integrates a Vision-Language Model (VLM) with each agent in a decentralized closed-loop manner under partial observability, improving both real-world applicability and scalability.

\subsection{LLM-based Multi-Agent System}
Based on the inspiring capabilities of LLMs, such as zero-shot planning and complex reasoning \cite{10.5555/3600270.3601883,10.5555/3666122.3667509,10.5555/3600270.3602070,Besta_Blach_Kubicek_Gerstenberger_Podstawski_Gianinazzi_Gajda_Lehmann_Niewiadomski_Nyczyk_Hoefler_2024}, embodied single-agent researches have demonstrated the effectiveness of LLMs in solving complex long-horizon tasks \cite{wang2024voyager,yao2023react,NEURIPS2023_1b44b878,ichter2022do,ma2024largelanguagemodelsplay,wang2024executablecodeactionselicit,tan2024cradleempoweringfoundationagents}. 
Despite significant advances in single-agent applications, developing real-world multi-agent systems with foundation models remains challenging, primarily due to the nature of decentralized control under partial observability in multi-agent settings \cite{10.5555/2283396.2283451}. 
Most prior efforts \cite{10610855,zhang2024buildingcooperativeembodiedagents,10.1609/aaai.v38i16.29710,nayak2025llamarlonghorizonplanningmultiagent,gong-etal-2024-mindagent} leverage a hierarchical framework with components like perception, communication, planning, execution, and memory to build multi-agent systems with collective behaviors. These approaches can be roughly classified into two groups. 1) Centralized plan: MindAgent \cite{gong-etal-2024-mindagent} adopts a centralized planning scheme with a pre-defined oracle in a fully observable multi-agent game. LLaMAR \cite{nayak2025llamarlonghorizonplanningmultiagent} employs LLMs to manage long-horizon tasks in partially observable environments without assumptions about access to perfect low-level policies. 2) Decentralized plan: ProAgent \cite{10.1609/aaai.v38i16.29710} introduces Theory of Mind (ToM), enabling agents to reason about others’ mental states. RoCo \cite{10610855} and CoELA \cite{zhang2024buildingcooperativeembodiedagents} assign separate LLMs to each embodied agent for collaboration with communication. However, RoCo and CoELA assume a skill library with a low-level heuristic controller, which is impractical in real-world applications. Moreover, RoCo's open-loop plan-and-execute paradigm fails to incorporate environmental feedback during decision-making. In contrast, our work does not assume any pre-defined low-level controller and generates code-based action through VLMs in a closed-loop manner.

\section{Preliminaries}
\label{background}
We model a fully cooperative multi-agent game with $N$ agents as a \textit{decentralized partially observable Markov decision process} (Dec-POMDP) \cite{10.5555/2967142}, which is formally defined as a tuple $\mathcal{G}=( \mathcal{N},\mathcal{S},\mathcal{O},\mathbb{O},\\ \mathcal{B}, \mathcal{A},\mathcal{T},\Omega,R,\gamma ,\rho_0 )$. $\mathcal{N} = \{ 1,\ldots,N \}$ is a set of agents, $s \in \mathcal{S}$ denotes the state of the environment and $\rho_0$ is the distribution of the initial state. $\mathcal{A} = \prod_{i=1}^{N}A^i$ is the joint action space, $\mathbb{O}=\prod_{i=1}^{N}O^i$ is the set of joint observations. At time step $t$, each agent $i$ receives an individual partial observation $o^i_t \in O^i$ given by the observation function $\mathcal{O} : (a_t, s_{t+1}) \mapsto  P(o_{t+1}|a_t, s_{t+1}) $ where $a_t, s_{t+1}$ and $o_{t+1}$ are the joint actions, states and joint observations respectively. Each agent $i$ uses a stochastic policy $\pi^{i}(a^i_t | h^i_t,\omega^i_t)$ conditioned on its action-observation history $h^i_t=(o^i_0, a^i_0, \ldots , o^i_{t-1}, a^i_{t-1})$ and a random seed $\omega^{i}_{t} \in \Omega_{t}$ to choose an action $a^i_t \in A^i$. A belief state $b_t$ is a probability distribution over states at time $t$, where $b_t \in \mathcal{B}$, and $\mathcal{B}$ is the space of all probability distributions over the state space. Actions $a_t$ drawn from joint policy $\pi(a_t|s_t, \omega_t)$ conditioned on state $s_t$ and joint random seed $\omega_t=(\omega^1_t, \ldots, \omega^N_t)$ change the state according to transition function $\mathcal{T} : (s_{t}, a^1_t, \ldots, a^N_t) \mapsto  P(s_{t+1}|s_{t}, a^1_t, \ldots, a^N_t) $. All agents share the same reward $r_t=R(s_{t}, a^1_t, \ldots, a^N_t)$ based on $s_t$ and $a_t$. $\gamma$ is the discount factor for future rewards. Agents try to maximize the expected total reward, $\mathcal{J}(\pi) = \mathbb{E}_{s_{0},a_0,\dots }[ \sum_{t=0}^{\infty}\gamma ^{t}r_t ]$, where $s_0 \sim \rho _0 (s_0), a_t \sim \pi (a_t|s_t,\omega_t)$.

\begin{figure}[t]
\centering
    \includegraphics[width=0.9\linewidth]{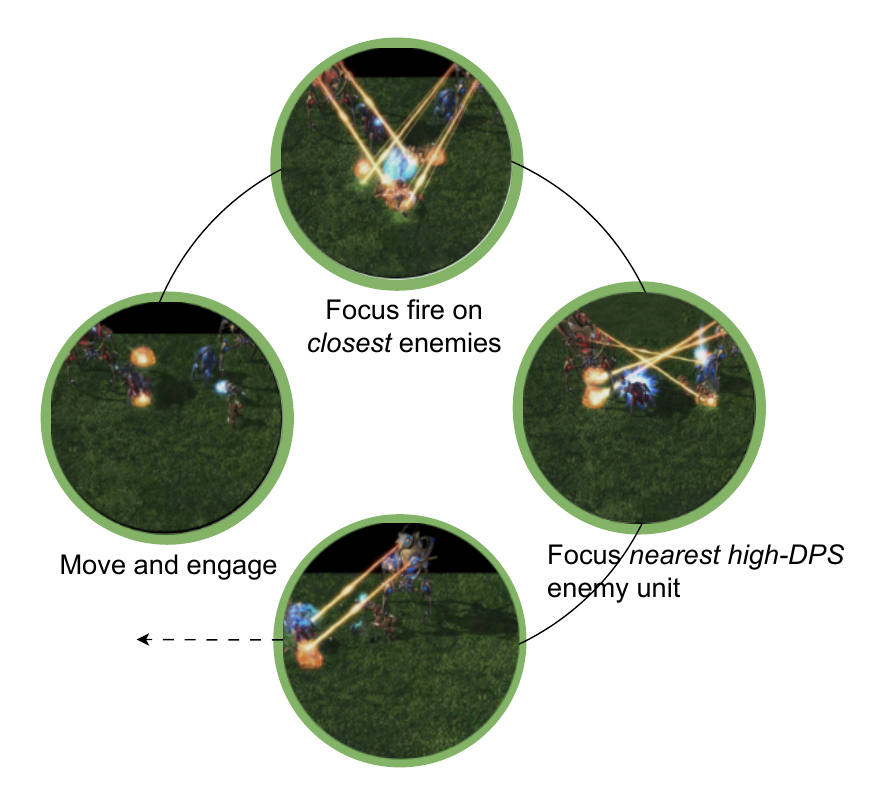}
    \caption{Visualization of COMPASS's dynamic task reasoning process in the StarCraft Multi-Agent Challenge (SMACv2) environment. The figure demonstrates how the VLM-based planner decomposes a complex final goal ("defeat all enemy units") into a sequence of concrete, executable sub-tasks that adapt to the changing battlefield conditions. This closed-loop task decomposition enables efficient coordination among multiple agents under partial observability, as each sub-task provides clear, actionable objectives that agents can execute while maintaining overall mission alignment.} 
    \Description{Visualization of COMPASS's dynamic task reasoning process in the StarCraft Multi-Agent Challenge (SMACv2) environment. The figure demonstrates how the VLM-based planner decomposes a complex final goal ("defeat all enemy units") into a sequence of concrete, executable sub-tasks that adapt to the changing battlefield conditions. This closed-loop task decomposition enables efficient coordination among multiple agents under partial observability, as each sub-task provides clear, actionable objectives that agents can execute while maintaining overall mission alignment.}
    \label{fig:task_reasoning}
\end{figure}

\section{Methods}
\label{methods}
COMPASS, illustrated in Figure \ref{fig:compass}, is a decentralized closed-loop framework for cooperative multi-agent systems that continuously incorporate environmental feedback for strategy refinement. The architecture comprises three core components: 1) a VLM-based closed-loop planner that iteratively perceives, reasons, reflects and acts to adaptively complete tasks (Sec. \ref{planner}); 2) an adaptive skill synthesis mechanism for generating executable codes tailored to proposed sub-tasks (Sec. \ref{skill-library}); and 3) a structured communication protocol that enables agents to share visible entity information under partial observability (Sec. \ref{communication}). The pseudo-code of COMPASS is shown in Appendix.

\subsection{VLM-based Closed-Loop Planner}
\label{planner}
Inspired by recent advances in cognitive architectures for autonomous systems \cite{tan2024cradleempoweringfoundationagents}, COMPASS implements a sophisticated modular planning framework that emulates key aspects of cognitive decision-making. The planner adopts a modular formulation, utilizing four specialized models: Perception, Task Reasoning, Self-Reflection, and Actor. Each model fulfills a distinct yet interconnected role in the decision-making process. The Perception model processes multi-modal inputs, integrating both visual and textual information to build comprehensive environmental understanding. The Task Reasoning model analyzes the perceived information to decompose complex objectives into manageable sub-tasks, ensuring systematic progress toward the final goal. The Self-Reflection model continuously evaluates task execution and outcome quality, enabling adaptive behavior refinement. The Actor model translates plans into actions by selecting and executing the most appropriate skills from the skill library. We next discuss the various components in detail:

\textbf{Perception} forms the foundation of COMPASS's decision-making capabilities by enabling robust multi-modal understanding of complex environments. Solving complex real-world tasks often involves data of multiple modalities \cite{10.1145/3613905.3651029}, each contributing unique and complementary information for decision-making. We leverage the VLMs' ability to fuse and analyze a broader spectrum of data, including text- and image-based environment feedback, to enable agents to sense the surrounding environment. The system's perception mechanism operates at two levels: direct observation processing and collaborative information synthesis. At the direct level, VLMs process raw inputs to extract meaningful features and relationships from both visual and textual data. At the collaborative level, COMPASS addresses the inherent challenge of partial observability in multi-agent systems through an innovative multi-hop communication protocol (detailed in the Structured Communication Protocol section) that enables agents to construct a more holistic understanding of their environment by sharing and aggregating observations. This dual-level perception architecture ensures that each agent maintains both detailed local awareness and broader contextual understanding, essential for effective decision-making in complex cooperative tasks.

\textbf{Task Reasoning} enables COMPASS to systematically approach complex cooperative challenges through collective task decomposition. Given a simple general final task in the cooperative multi-agent setting, e.g., "defeat all enemy units", in order to complete the task more efficiently, agents are required to decompose it into multiple sub-tasks and figure out the right one to focus on, while considering alignment among others (See Figure \ref{fig:task_reasoning}). COMPASS harnesses the power of VLMs to analyze high-level task instructions in conjunction with environmental feedback and team member objectives to generate tractable sub-tasks that collectively advance the overall mission. As agents act under stochastic, partially observable environments, the task reasoning model continuously adapts its plans, proposing and refining sub-tasks based on emerging situations and progress assessment. This dynamic approach enables COMPASS to maintain strategic coherence while adjusting tactical decisions in response to changing circumstances.

\textbf{Actor} serves as the critical bridge between high-level reasoning and concrete action execution. Building upon recent advances in code-writing language models for embodied control \cite{liang2023codepolicieslanguagemodel,wang2024executablecodeactionselicit}, the Actor leverages the skill library by first identifying relevant skills for the proposed sub-task, then synthesizes perception and self-reflection inputs to select the optimal skill for execution. This streamlined approach ensures efficient skill selection while maintaining task alignment.

\begin{figure}[t]
\centering
    \includegraphics[width=0.9\linewidth]{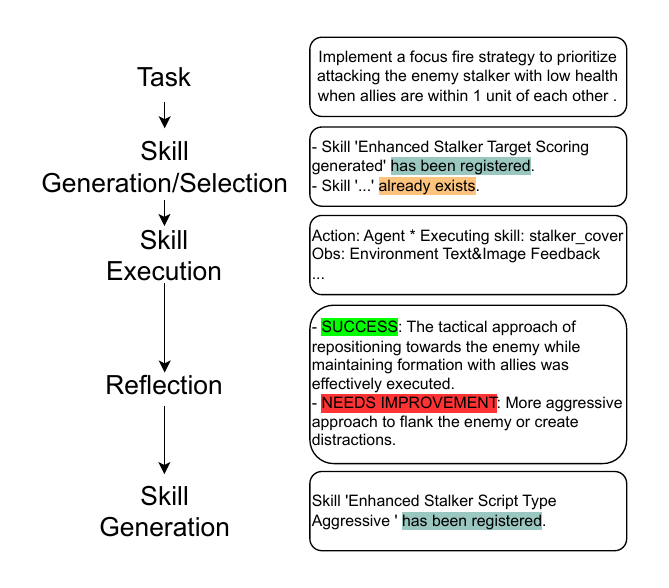}
    \caption{Dynamic skill evolution through self-reflection. COMPASS continuously refines its tactical capabilities by analyzing the performance of executed skills. Here, feedback on the `stalker\_cover' skill reveals a need for more aggression. This insight prompts the immediate generation and registration of a new, specialized skill (`Enhanced Stalker Script Type Aggressive'), expanding the agent's behavioral repertoire.} 
    \Description{Dynamic skill evolution through self-reflection. COMPASS continuously refines its tactical capabilities by analyzing the performance of executed skills. Here, feedback on the `stalker\_cover' skill reveals a need for more aggression. This insight prompts the immediate generation and registration of a new, specialized skill (`Enhanced Stalker Script Type Aggressive'), expanding the agent's behavioral repertoire.}
    \label{fig:self_reflection}
\end{figure}

\begin{figure*}[t]
\centering
\includegraphics[width=\linewidth]{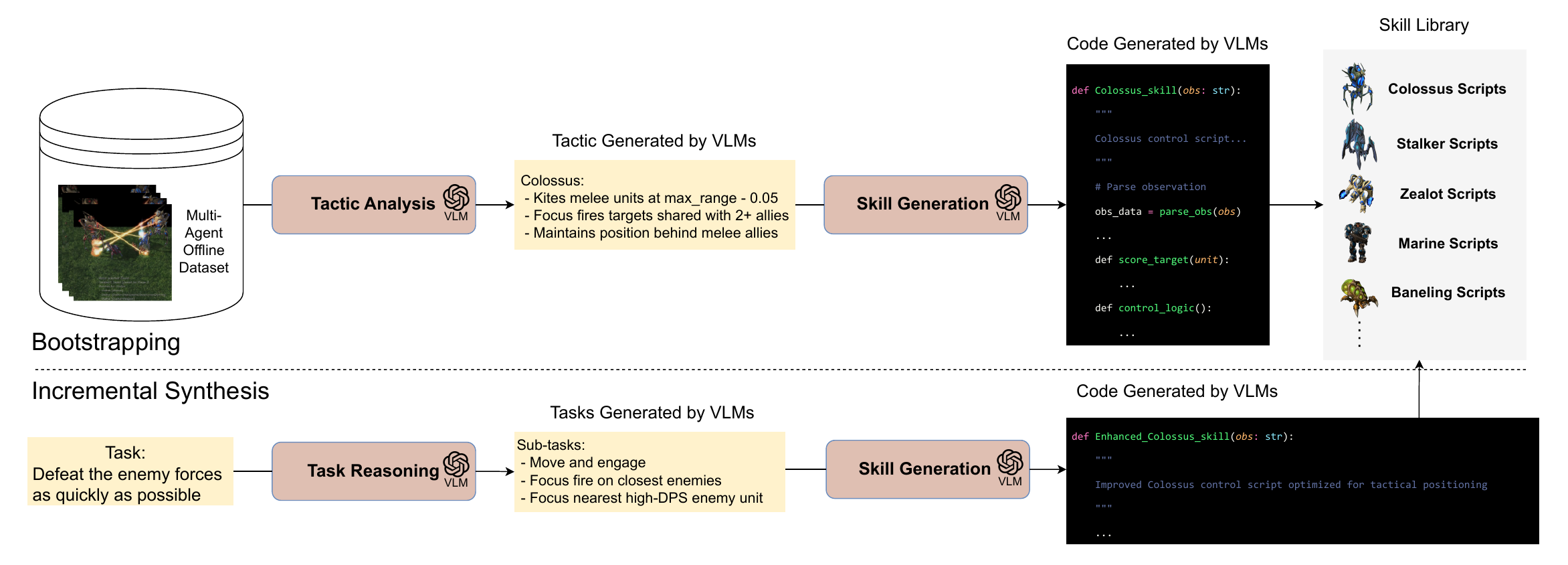}
\caption{Overview of Adaptive Skill Synthesis. VLMs perform (Top) Bootstrapping by analyzing offline data for initial Tactic Analysis and Skill Generation into a Skill Library. (Bottom) Incremental Synthesis uses Task Reasoning to dynamically generate or enhance code-based skills, evolving the library for new tasks. The skills follow a structured decision-making pipeline with two core components: \textit{score\_target(unit)} for dynamic target prioritization and \textit{control\_logic()} for coordinating behavior. Textual observations are parsed into structured data (obs\_data), mapping raw text to attributes, e.g., ``Can move North: yes" to can\_move={`north': True}.}
\Description{Overview of Adaptive Skill Synthesis. VLMs perform (Top) Bootstrapping by analyzing offline data for initial Tactic Analysis and Skill Generation into a Skill Library. (Bottom) Incremental Synthesis uses Task Reasoning to dynamically generate or enhance code-based skills, evolving the library for new tasks. The skills follow a structured decision-making pipeline with two core components: \textit{score\_target(unit)} for dynamic target prioritization and \textit{control\_logic()} for coordinating behavior. Textual observations are parsed into structured data (obs\_data), mapping raw text to attributes, e.g., ``Can move North: yes" to can\_move={`north': True}.}
\label{fig:adaptive_skill}
\end{figure*}

\textbf{Self-Reflection} enables COMPASS to continuously evaluate and refine its decision-making processes through systematic performance analysis (See Figure \ref{fig:self_reflection}). COMPASS instantiates the Self-Reflection model as a VLM which takes a sequence of visual results from the last skill execution with corresponding descriptions as input to assess the quality of the decision produced by the Actor and whether the task was completed. Additionally, we also request the VLM to generate verbal self-reflections to provide valuable feedback on the completion of the task.

\subsection{Adaptive Skill Synthesis}
\label{skill-library}
COMPASS employs a dynamic skill library that maintains and evolves a collection of executable behaviors. Each skill is represented as an executable Python function with comprehensive documentation describing its functionality and corresponding embedding that enables semantic retrieval. This skill library undergoes continuous refinement through two complementary mechanisms (Figure \ref{fig:adaptive_skill}): incremental synthesis, where new skills are generated and existing ones are refined during task execution, and demonstration-based bootstrapping, which initializes the library with behaviors extracted from expert demonstrations.

\textbf{Incremental Synthesis}
With the Task Reasoning component consistently proposing sub-tasks, COMPASS first attempts to retrieve relevant skills from the library using semantic similarity between the sub-task description and skill documentation embeddings. If no suitable skill exists, or if existing skills prove inadequate, the VLM generates a new Python script specifically tailored to the sub-task.

\begin{figure}[t]
\centering
    \includegraphics[width=0.9\linewidth]{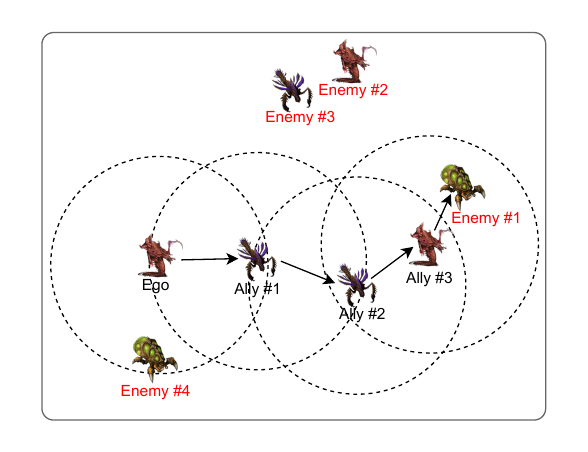}
    \caption{Illustration of COMPASS's structured multi-hop communication protocol that enables efficient information sharing under partial observability. The figure demonstrates how information about Enemy \#1 propagates to the Ego agent through a chain of allied units (Ally \#1, \#2, \#3), despite Enemy \#1 being outside Ego's sight range. Each dashed circle represents an agent's local observation field, while arrows indicate the flow of entity-based information sharing. This mechanism enables agents to build a more holistic understanding of the environment by propagating critical information (e.g., enemy positions, status) through intermediate allies, effectively addressing the partial observability challenge in decentralized multi-agent systems.} 
    \Description{Illustration of COMPASS's structured multi-hop communication protocol that enables efficient information sharing under partial observability. The figure demonstrates how information about Enemy \#1 propagates to the Ego agent through a chain of allied units (Ally \#1, \#2, \#3), despite Enemy \#1 being outside Ego's sight range. Each dashed circle represents an agent's local observation field, while arrows indicate the flow of entity-based information sharing. This mechanism enables agents to build a more holistic understanding of the environment by propagating critical information (e.g., enemy positions, status) through intermediate allies, effectively addressing the partial observability challenge in decentralized multi-agent systems.}
    \label{fig:multi-hop}
\end{figure}

\textbf{Bootstrapping}
However, developing the skill library from scratch requires extensive interactions with environments, which potentially leads to inefficient learning in the early stages. Inspired by offline MARL approaches \cite{meng2023offline,zhang2022discovering,zhu2025madiffofflinemultiagentlearning}, which leverage pre-collected datasets to enhance sample efficiency, we leverage MAPPO as the behavior policy to collect experiences, which are recorded as video sequences. The VLMs then analyze these demonstrations through a multi-stage process: first identifying key strategic patterns and behavioral primitives, then translating these patterns into executable Python functions with appropriate documentation. This initialization methodology establishes a foundational set of validated skills, substantially reducing the exploration overhead typically required for discovering effective behaviors. The resulting baseline skill library enables efficient task execution from the onset while maintaining the flexibility to evolve through incremental synthesis.

\begin{figure*}[ht!]
\centering
\includegraphics[width=0.85\linewidth]{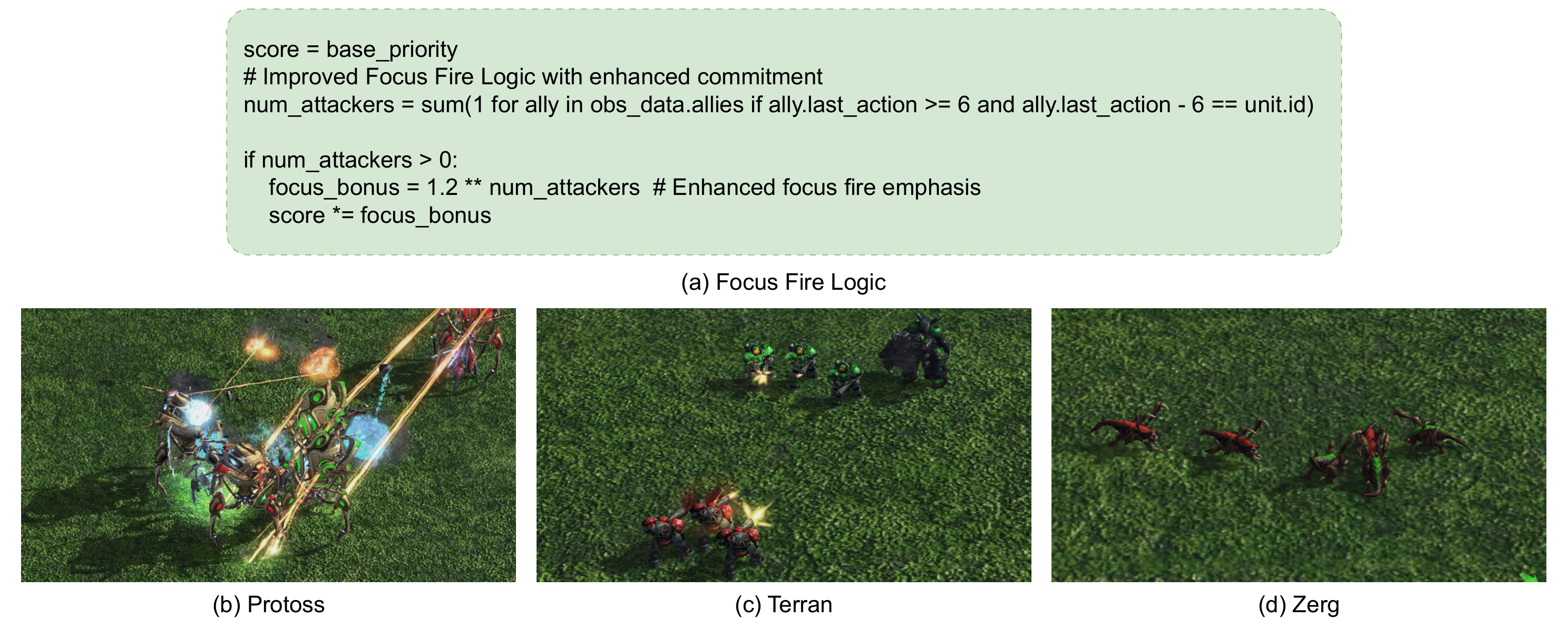} 
\caption{Focus Fire Logic Implementation. (a) VLM-generated Python code snippet implementing dynamic focus fire logic. The code prioritizes enemy units based on the number of allied attackers, scaling the attack bonus exponentially. (b–d) Visualizations of focus fire execution across Protoss, Terran, and Zerg.}
\Description{Focus Fire Logic Implementation. (a) VLM-generated Python code snippet implementing dynamic focus fire logic. The code prioritizes enemy units based on the number of allied attackers, scaling the attack bonus exponentially. (b–d) Visualizations of focus fire execution across Protoss, Terran, and Zerg.}
\label{fig:smac_focus_fire}
\end{figure*}

\textbf{Skill Analysis.} We now analyze COMPASS's capability to synthesize and execute diverse tactical behaviors. COMPASS develops four key tactical patterns: (1) An exponentially-scaled focus fire implementation that coordinates multiple units' target selection based on allied attacker density (Figure \ref{fig:smac_focus_fire}), (2) A position-aware kiting mechanism that maintains optimal engagement ranges while managing unit positioning relative to threats (Figure \ref{fig:kitting}), (3) A formation-based isolation tactic that enables systematic target elimination through coordinated unit movements (Figure \ref{fig:isolate}), and (4) An area-of-effect (AOE) tactic that maximizes splash damage through cluster density calculation (Figure \ref{fig:baneling}). These synthesized skills exhibit clear strategic intent while maintaining interpretability.

\subsection{Structured Communication Protocol}
\label{communication}
To facilitate effective collaboration under partial observability, recent LLM-based multi-agent work \cite{10.5555/3666122.3668386,zhang2024buildingcooperativeembodiedagents} employs conversational framework with unconstrained communication protocol. However, while natural language offers flexibility, unrestricted communication can lead to potential hallucinations caused by ambiguous or irrelevant messages between agents. Drawing from advances in structured communication frameworks \cite{hong2024metagpt} and entity-based MARL \cite{pmlr-v139-iqbal21a,pmlr-v202-ding23d}, COMPASS implements a hierarchical communication protocol that focuses on efficient entity-based information sharing and multi-hop propagation (Figure \ref{fig:multi-hop}). Each agent maintains an observation buffer containing information about entities in its field of view. At each timestep, agents share their local observations, which are then aggregated into a global entity memory accessible to all. COMPASS employs a multi-hop communication mechanism to propagate information about distant entities, enabling agents to build a more holistic observation of the environment by leveraging the collective knowledge of the team.

\section{Experiments}
\label{experiments}
We conducted a comprehensive experimental evaluation of COMPASS to assess its performance and capabilities in complex multi-agent scenarios. Our evaluation focused on the improved StarCraft Multi-Agent Challenge (SMACv2) \cite{10.5555/3666122.3667756}, which provides an ideal testbed for examining cooperative behavior under partial observability and stochasticity. Through systematic experimentation, we investigated two fundamental questions: (1) How does COMPASS perform compared to state-of-the-art MARL methods? (2) What are the individual contributions of each component in COMPASS? Experiments utilize both open-source (Qwen2-VL-72B) and closed-source VLMs (GPT-4o-mini, Claude-3-Haiku), with Jina AI embeddings for skill retrieval. All results are averaged over 5 seeds to account for environmental stochasticity. Token usage is approximately 0.4 million per episode.

\subsection{Experimental Setup}
\textbf{Scenarios} Our evaluation scenarios span three distinct race matchups (Protoss, Terran, and Zerg) and two categories (symmetric and asymmetric), as detailed in Appendix. The symmetric scenarios (5v5) test coordination in balanced engagements, while asymmetric scenarios (5v6) evaluate adaptation to numerical disadvantages. Each race combination presents unique tactical challenges due to different unit abilities and constraints. We followed the setting p=0 in the SMACv2 original paper (i.e., prob\_obs\_enemy: 0.0 in the .yaml file), meaning that only the first agent to initially spot a specific enemy unit can continue observing it, introducing the \textit{Extended Partial Observability Challenge}, which baselines struggled with.
\begin{table}[ht]
\caption{Evaluation scenarios spanning three SMACv2 race matchups under symmetric equal-force and asymmetric outnumbered configurations.}
\label{tab:smacv2}
\begin{center}
\begin{small}
\begin{sc}
\begin{tabular}{lcc}
\toprule
Task & Scenarios & Categories\\
\midrule
Protoss & protoss 5 vs 5 & symmetric\\
  & protoss 5 vs 6 & asymmetric\\
Terran & terran 5 vs 5 & symmetric\\
  & terran 5 vs 6 & asymmetric\\
Zerg & zerg 5 vs 5 & symmetric\\
  & zerg 5 vs 6 & asymmetric\\
\bottomrule
\end{tabular}
\end{sc}
\end{small}
\end{center}
\end{table}

\textbf{Baselines} We compared COMPASS against the state-of-the-art MARL algorithms representing both value-based and policy-gradient approaches:
\begin{itemize}
    \item Value-Based Methods: QMIX \cite{JMLR:v21:20-081} uses a mixing network architecture to decompose joint action-values while maintaining monotonicity constraints.
    \item Policy Gradient Methods: MAPPO \cite{NEURIPS2022_9c1535a0} extends PPO to multi-agent settings with the CTDE paradigm. HAPPO \cite{kuba2022trustregionpolicyoptimisation} performs sequential policy updates by utilizing other agents' newest policy under the CTDE framework and provably obtains the monotonic policy improvement guarantee. HASAC \cite{liu2024maximum} combines the maximum entropy framework with trust region optimization to enhance exploration and coordination. MAT\cite{NEURIPS2022_69413f87} models the multi-agent decision process as a sequence-to-sequence generation problem with a powerful transformer architecture.
    \item Communication-based Methods: CommFormer \cite{hu2024learning} is a method for learning optimal communication graphs in multi-agent systems using attention.
    \item Offline Methods: Oryx \cite{formanekoryx} is a state-of-the-art offline MARL algorithm, addressing extrapolation error and miscoordination.
    \item LLM-based Methods: LLM-SMAC \cite{deng2024newapproachsolvingsmac} generates decision tree code using Large Language Models (GPT-4o-mini) to solve the SMAC task. We modified its codebase for SMACv2, but the resulting win rates were 0\% across all settings. Given this performance, we ignore this baseline in our final comparison.
\end{itemize}

\textbf{Datasets} To enable effective bootstrapping of the skill library, we constructed a comprehensive demonstration dataset capturing diverse multi-agent strategies and interactions. We employed MAPPO with original hyper-parameters as our behavior policy for data collection, leveraging its strong performance in cooperative multi-agent tasks. Our final dataset comprises over 300 complete game episodes, each recorded as a video sequence capturing the full state-action trajectory. These demonstrations span all symmetric scenario types described in Appendix.

\subsection{Main Results}
\begin{table*}[t]
\centering
\caption{Comparative performance of COMPASS (with three VLM variants: G-4o=GPT-4o-mini, C-Hk=Claude-3-Haiku, Q2-VL=Qwen2-VL-72B) and state-of-the-art baselines on SMACv2. Median win rates (\%) and standard deviations (subscripts) are reported across Protoss, Terran, and Zerg scenarios in symmetric (5v5) and asymmetric (5v6) categories. Results are averaged over 5 seeds. Bold values denote the best performance in each scenario. N/A* indicates no datasets available in these settings.}
\begin{tabular}{llccccccc}
\toprule
& & \multicolumn{2}{c}{\textbf{Protoss}} & \multicolumn{2}{c}{\textbf{Terran}} & \multicolumn{2}{c}{\textbf{Zerg}} \\
\cmidrule(lr){3-4} \cmidrule(lr){5-6} \cmidrule(lr){7-8}
\textbf{Method} & \textbf{Type} & 5v5 & 5v6 & 5v5 & 5v6 & 5v5 & 5v6 \\
\midrule
\multicolumn{8}{l}{\textit{Online MARL}} \\
QMIX & Online & $0.27_{0.03}$ & $0.01_{0.01}$ & $0.38_{0.04}$ & $0.06_{0.02}$ & $0.21_{0.01}$ & $\mathbf{0.18_{0.03}}$ \\
MAPPO & Online & $0.32_{0.07}$ & $0.04_{0.04}$ & $0.36_{0.10}$ & $0.07_{0.06}$ & $0.27_{0.04}$ & $0.13_{0.09}$ \\
HAPPO & Online & $0.34_{0.07}$ & $0.02_{0.03}$ & $0.35_{0.10}$ & $0.01_{0.03}$ & $0.20_{0.11}$ & $0.09_{0.02}$ \\
HASAC & Online & $0.20_{0.08}$ & $0.01_{0.02}$ & $0.29_{0.01}$ & $0.05_{0.02}$ & $0.24_{0.07}$ & $0.08_{0.05}$ \\
MAT & Online & $0.39_{0.03}$ & $0.04_{0.04}$ & $0.36_{0.11}$ & $0.05_{0.01}$ & $0.32_{0.06}$ & $0.11_{0.08}$ \\
CommFormer & Communication & $0.39_{0.16}$ & $0.02_{0.01}$ & $0.30_{0.09}$ & $0.03_{0.01}$ & $\mathbf{0.39_{0.10}}$ & $0.16_{0.01}$ \\
\midrule
\multicolumn{8}{l}{\textit{Offline MARL}} \\
Oryx & Offline & N/A* & N/A* & $0.18_{0.04}$ & N/A* & $0.10_{0.06}$ & N/A* \\
\midrule
\multicolumn{8}{l}{\textit{VLM-based (Ours)}} \\
COMPASS (G-4o) & VLM & $\mathbf{0.57_{0.08}}$ & $\mathbf{0.08_{0.04}}$ & $\mathbf{0.39_{0.01}}$ & $\mathbf{0.10_{0.03}}$ & $0.16_{0.07}$ & $0.03_{0.01}$ \\
COMPASS (C-Hk) & VLM & $0.49_{0.06}$ & $0.06_{0.05}$ & $0.38_{0.05}$ & $\mathbf{0.10_{0.01}}$ & $0.18_{0.02}$ & $0.04_{0.01}$ \\
COMPASS (Q2-VL) & VLM & $0.45_{0.04}$ & $0.06_{0.03}$ & $0.31_{0.02}$ & $0.06_{0.03}$ & $0.14_{0.03}$ & $0.02_{0.01}$ \\
\bottomrule
\end{tabular}
\label{tab:quantitative_results}
\end{table*}

\textbf{Performance.} Table \ref{tab:quantitative_results} reveals substantial performance gains for COMPASS in SMACv2, with the clearest advantages emerging in Protoss scenarios. Using GPT-4o-mini, COMPASS achieves a 57\% win rate in symmetric Protoss engagements, exceeding QMIX by 30 percentage points, MAPPO by 25 points, and HAPPO by 23 points. The margin over MAT and CommFormer remains significant at 18 percentage points despite their stronger baseline performance.

Performance stratifies sharply across race matchups. Terran scenarios yield a 39\% win rate, positioning COMPASS marginally above all baselines. Zerg scenarios expose a critical limitation: COMPASS attains only 16\% win rate, falling below CommFormer at 39\%. This disparity stems directly from Zerg unit mechanics. Banelings and Zerglings require continuous micromanagement at sub-second timescales due to their melee attack ranges and swarm coordination demands, while COMPASS queries the VLM every 10 to 20 timesteps. The mismatch between decision frequency and tactical requirements undermines combat effectiveness.

Asymmetric 5v6 scenarios test adaptation under numerical disadvantage. COMPASS maintains leads over MARL baselines in Protoss and Terran configurations, reaching 8\% and 10\% win rates respectively where most baselines achieve 1\% to 7\%. This suggests the skill library and structured communication enable tactical compensation for unit deficits. VLM choice affects outcomes moderately: GPT-4o-mini consistently outperforms Claude-3-Haiku and Qwen2-VL-72B by 2 to 8 percentage points across most scenarios, though all three variants follow the same performance ranking across races.

The Zerg 5v6 asymmetric case merits attention. COMPASS achieves only 2\% to 4\% win rate compared to QMIX at 18\% and CommFormer at 16\%. QMIX benefits from dense value function updates that capture rapid state changes, while COMPASS relies on periodic high-level replanning. When both numerical disadvantage and high-frequency control demands coincide, the VLM-based approach degrades substantially.

\subsection{Ablation Studies}
\begin{table}
\centering
\caption{Win rates achieved by the bootstrapped skill library alone, without incremental synthesis during deployment. Skills were extracted from MAPPO demonstration trajectories and evaluated directly on SMACv2 test episodes.}
\begin{tabular}{lccc}
\hline
& PROTOSS & TERRAN & ZERG \\
\hline
5V5 & $0.35_{0.06}$ & $0.24_{0.04}$ & $0.06_{0.01}$ \\
5V6 & $0.04_{0.05}$ & $0.06_{0.02}$ & $0.02_{0.03}$ \\
\hline
\end{tabular}
\label{tab:initialized}
\end{table}

\textbf{Skill Initialization} To evaluate the impact of our skill initialization, we analyze the performance of COMPASS using only the initialized skill library derived from expert demonstrations. The results in Table \ref{tab:initialized} demonstrate that skill initialization alone achieves non-trivial performance across different scenarios, particularly in symmetric matchups. Moreover, the gap between initialized skills and COMPASS underscores the necessity of incremental skill synthesis. A script example for skill initialization is in Appendix.

\textbf{Communication} To demonstrate the critical role of communication, we evaluated COMPASS on Protoss 5v5 under the \textit{Extended Partial Observability} setting, using only local information without multi-hop propagation. The resulting win rate with GPT-4o-mini decreased to $0.06_{0.04}$, a significant drop from $0.57$ with full communication. This degradation occurs because the Extended Partial Observability setting restricts direct enemy visibility to the first agent that initially spots it. The VLMs generated control logic heavily relies on the presence of enemies in the local observation to determine engagement and targeting. Without communication relaying enemy positions, agents other than the discoverer cannot `see' enemies known to teammates, even if within attack range. Consequently, their control logic frequently defaults to `no enemy' behaviors, such as moving towards allies or executing random default actions, preventing effective target engagement and coordinated attacks, thus drastically reducing combat effectiveness and the overall win rate. We study the effect of a communication fault. As shown in the Table \ref{tab:comm_robustness}, the system degrades under moderate packet loss and suffers a drastic drop at >50\% loss.

\begin{table}[t]
\centering
\caption{Communication protocol resilience on Protoss 5v5 scenarios. Top: win rates increase with propagation depth as agents access information beyond direct observation range. Bottom: performance degrades gracefully under moderate packet loss but collapses beyond 50\% loss when multi-hop propagation fails.}
\begin{tabular}{lcccc}
\toprule
Setting & Specification & Win Rate \\
\midrule
Hop Count & 1-hop & $0.46_{0.19}$ \\
 & 2-hop & $0.54_{0.06}$ \\
 & 3-hop & $\mathbf{0.57_{0.08}}$ \\
\midrule
Packet Loss & 20\% & $\mathbf{0.32_{0.03}}$ \\
 & 50\% & $0.12_{0.02}$ \\
 & 80\% & $0.07_{0.05}$ \\
 & 100\% & $0.06_{0.04}$ \\
\bottomrule
\end{tabular}
\label{tab:comm_robustness}
\end{table}

\textbf{VLM call frequency} We study the impact of VLM call frequency. As shown in the Table \ref{tab:vlm_freq}, at higher call frequencies, the performance remains similar but incurs greater cost, while lower frequencies fail to keep up with the dynamics of the battlefield.

\begin{table}[t]
\centering
\caption{VLM query frequency ablation on Protoss 5v5 scenarios with average episode length of 60 steps. Performance remains stable between 10-step and 20-step intervals but degrades at 40-step intervals when replanning cannot track battlefield dynamics.}
\begin{tabular}{lcc}
\toprule
Frequency & Win Rate \\
\midrule
Every 10 steps & $0.56_{0.05}$ \\
Every 20 steps & $\mathbf{0.57_{0.08}}$ \\
Every 40 steps & $0.40_{0.08}$ \\
\bottomrule
\end{tabular}
\label{tab:vlm_freq}
\end{table}

\textbf{Self Reflection} In order to show the effectiveness of self-reflection, we evaluate the performance of COMPASS w/o self-reflection on protoss 5 vs 5. Removing the module leads to a drop of 10\% ($0.47_{0.04}$).

\textbf{Visual information} We tested visual information contribution by omitting image inputs. This forces agents to rely solely on textual information, eliminating visual grounding for spatial understanding. Performance drops 10\% without visual input. Analysis of VLM outputs reveals that absent visual information, the VLM must infer map boundaries from textual cues such as `West (unavailable movement)' that encode action constraints rather than explicit spatial geometry. Visual input enables direct perception of these spatial details from the image. The resulting degradation in spatial awareness produces suboptimal tactical decisions for movement and positioning.

\section{Conclusion}
\label{conclusion}
COMPASS demonstrates that vision-language models can generate interpretable tactical behaviors for cooperative multi-agent control through skill synthesis and structured communication under partial observability. The framework achieves a 57\% win rate on symmetric Protoss scenarios in SMACv2, exceeding the QMIX baseline by 30 percentage points and outperforming all tested MARL methods. However, performance collapses on Zerg scenarios to 16\% win rate, falling below several baselines including CommFormer at 39\%. This failure exposes a fundamental constraint: VLM query intervals of 10 to 20 timesteps cannot provide the continuous low-level control required for melee swarm units with sub-second tactical windows. The results indicate that VLM-based planning offers advantages for scenarios where strategic coordination dominates and where actions retain validity across multiple timesteps, but the approach remains unsuitable for domains requiring rapid reactive control. Future work must either reduce VLM inference latency by orders of magnitude or develop hybrid architectures that delegate high-frequency decisions to learned reactive policies while reserving VLMs for strategic supervision.



\bibliographystyle{ACM-Reference-Format} 
\bibliography{sample}

@book{10.5555/2967142,
author = {Oliehoek, Frans A. and Amato, Christopher},
title = {A Concise Introduction to Decentralized POMDPs},
year = {2016},
isbn = {3319289276},
publisher = {Springer Publishing Company, Incorporated},
edition = {1st}
}

@misc{vinyals2017starcraftiinewchallenge,
      title={StarCraft II: A New Challenge for Reinforcement Learning}, 
      author={Oriol Vinyals and Timo Ewalds and Sergey Bartunov and Petko Georgiev and Alexander Sasha Vezhnevets and Michelle Yeo and Alireza Makhzani and Heinrich Küttler and John Agapiou and Julian Schrittwieser and John Quan and Stephen Gaffney and Stig Petersen and Karen Simonyan and Tom Schaul and Hado van Hasselt and David Silver and Timothy Lillicrap and Kevin Calderone and Paul Keet and Anthony Brunasso and David Lawrence and Anders Ekermo and Jacob Repp and Rodney Tsing},
      year={2017},
      eprint={1708.04782},
      archivePrefix={arXiv},
      primaryClass={cs.LG},
      url={https://arxiv.org/abs/1708.04782}, 
}

@inproceedings{10.5555/3306127.3332052,
author = {Samvelyan, Mikayel and Rashid, Tabish and Schroeder de Witt, Christian and Farquhar, Gregory and Nardelli, Nantas and Rudner, Tim G. J. and Hung, Chia-Man and Torr, Philip H. S. and Foerster, Jakob and Whiteson, Shimon},
title = {The StarCraft Multi-Agent Challenge},
year = {2019},
isbn = {9781450363099},
publisher = {International Foundation for Autonomous Agents and Multiagent Systems},
address = {Richland, SC},
booktitle = {Proceedings of the 18th International Conference on Autonomous Agents and MultiAgent Systems},
pages = {2186–2188},
numpages = {3},
location = {Montreal QC, Canada},
series = {AAMAS '19}
}

@inproceedings{10.5555/3666122.3667756,
author = {Ellis, Benjamin and Cook, Jonathan and Moalla, Skander and Samvelyan, Mikayel and Sun, Mingfei and Mahajan, Anuj and Foerster, Jakob N. and Whiteson, Shimon},
title = {SMACv2: an improved benchmark for cooperative multi-agent reinforcement learning},
year = {2024},
publisher = {Curran Associates Inc.},
address = {Red Hook, NY, USA},
booktitle = {Proceedings of the 37th International Conference on Neural Information Processing Systems},
articleno = {1634},
numpages = {27},
location = {New Orleans, LA, USA},
series = {NIPS '23}
}

@article{JMLR:v21:20-081,
  author  = {Tabish Rashid and Mikayel Samvelyan and Christian Schroeder de Witt and Gregory Farquhar and Jakob Foerster and Shimon Whiteson},
  title   = {Monotonic Value Function Factorisation for Deep Multi-Agent Reinforcement Learning},
  journal = {Journal of Machine Learning Research},
  year    = {2020},
  volume  = {21},
  number  = {178},
  pages   = {1--51},
  url     = {http://jmlr.org/papers/v21/20-081.html}
}

@inproceedings{NEURIPS2022_9c1535a0,
 author = {Yu, Chao and Velu, Akash and Vinitsky, Eugene and Gao, Jiaxuan and Wang, Yu and Bayen, Alexandre and WU, YI},
 booktitle = {Advances in Neural Information Processing Systems},
 editor = {S. Koyejo and S. Mohamed and A. Agarwal and D. Belgrave and K. Cho and A. Oh},
 pages = {24611--24624},
 publisher = {Curran Associates, Inc.},
 title = {The Surprising Effectiveness of PPO in Cooperative Multi-Agent Games},
 url = {https://proceedings.neurips.cc/paper_files/paper/2022/file/9c1535a02f0ce079433344e14d910597-Paper-Datasets_and_Benchmarks.pdf},
 volume = {35},
 year = {2022}
}

@misc{kuba2022trustregionpolicyoptimisation,
      title={Trust Region Policy Optimisation in Multi-Agent Reinforcement Learning}, 
      author={Jakub Grudzien Kuba and Ruiqing Chen and Muning Wen and Ying Wen and Fanglei Sun and Jun Wang and Yaodong Yang},
      year={2022},
      eprint={2109.11251},
      archivePrefix={arXiv},
      primaryClass={cs.AI},
      url={https://arxiv.org/abs/2109.11251}, 
}

@inproceedings{
liu2024maximum,
title={Maximum Entropy Heterogeneous-Agent Reinforcement Learning},
author={Jiarong Liu and Yifan Zhong and Siyi Hu and Haobo Fu and QIANG FU and Xiaojun Chang and Yaodong Yang},
booktitle={The Twelfth International Conference on Learning Representations},
year={2024},
url={https://openreview.net/forum?id=tmqOhBC4a5}
}

@article{Li_Liu_Zhang_Wei_Niu_Yang_Liu_Ouyang_2023, title={ACE: Cooperative Multi-Agent Q-learning with Bidirectional Action-Dependency}, volume={37}, url={https://ojs.aaai.org/index.php/AAAI/article/view/26028}, DOI={10.1609/aaai.v37i7.26028}, abstractNote={Multi-agent reinforcement learning (MARL) suffers from the non-stationarity problem, which is the ever-changing targets at every iteration when multiple agents update their policies at the same time. Starting from first principle, in this paper, we manage to solve the non-stationarity problem by proposing bidirectional action-dependent Q-learning (ACE). Central to the development of ACE is the sequential decision making process wherein only one agent is allowed to take action at one time. Within this process, each agent maximizes its value function given the actions taken by the preceding agents at the inference stage. In the learning phase, each agent minimizes the TD error that is dependent on how the subsequent agents have reacted to their chosen action. Given the design of bidirectional dependency, ACE effectively turns a multi-agent MDP into a single-agent MDP. We implement the ACE framework by identifying the proper network representation to formulate the action dependency, so that the sequential decision process is computed implicitly in one forward pass. To validate ACE, we compare it with strong baselines on two MARL benchmarks. Empirical experiments demonstrate that ACE outperforms the state-of-the-art algorithms on Google Research Football and StarCraft Multi-Agent Challenge by a large margin. In particular, on SMAC tasks, ACE achieves 100% success rate on almost all the hard and super hard maps. We further study extensive research problems regarding ACE, including extension, generalization and practicability.}, number={7}, journal={Proceedings of the AAAI Conference on Artificial Intelligence}, author={Li, Chuming and Liu, Jie and Zhang, Yinmin and Wei, Yuhong and Niu, Yazhe and Yang, Yaodong and Liu, Yu and Ouyang, Wanli}, year={2023}, month={Jun.}, pages={8536-8544} }

@inproceedings{NEURIPS2022_69413f87,
 author = {Wen, Muning and Kuba, Jakub and Lin, Runji and Zhang, Weinan and Wen, Ying and Wang, Jun and Yang, Yaodong},
 booktitle = {Advances in Neural Information Processing Systems},
 editor = {S. Koyejo and S. Mohamed and A. Agarwal and D. Belgrave and K. Cho and A. Oh},
 pages = {16509--16521},
 publisher = {Curran Associates, Inc.},
 title = {Multi-Agent Reinforcement Learning is a Sequence Modeling Problem},
 url = {https://proceedings.neurips.cc/paper_files/paper/2022/file/69413f87e5a34897cd010ca698097d0a-Paper-Conference.pdf},
 volume = {35},
 year = {2022}
}

@inproceedings{
hu2024learning,
title={Learning Multi-Agent Communication from Graph Modeling Perspective},
author={Shengchao Hu and Li Shen and Ya Zhang and Dacheng Tao},
booktitle={The Twelfth International Conference on Learning Representations},
year={2024},
url={https://openreview.net/forum?id=Qox9rO0kN0}
}

@inproceedings{
na2024efficient,
title={Efficient Episodic Memory Utilization of Cooperative Multi-Agent Reinforcement Learning},
author={Hyungho Na and Yunkyeong Seo and Il-chul Moon},
booktitle={The Twelfth International Conference on Learning Representations},
year={2024},
url={https://openreview.net/forum?id=LjivA1SLZ6}
}

@article{meng2023offline,
  title={Offline pre-trained multi-agent decision transformer},
  author={Meng, Linghui and Wen, Muning and Le, Chenyang and Li, Xiyun and Xing, Dengpeng and Zhang, Weinan and Wen, Ying and Zhang, Haifeng and Wang, Jun and Yang, Yaodong and others},
  journal={Machine Intelligence Research},
  volume={20},
  number={2},
  pages={233--248},
  year={2023},
  publisher={Springer}
}

@inproceedings{zhang2022discovering,
  title={Discovering generalizable multi-agent coordination skills from multi-task offline data},
  author={Zhang, Fuxiang and Jia, Chengxing and Li, Yi-Chen and Yuan, Lei and Yu, Yang and Zhang, Zongzhang},
  booktitle={The Eleventh International Conference on Learning Representations},
  year={2022}
}

@misc{zhu2025madiffofflinemultiagentlearning,
      title={MADiff: Offline Multi-agent Learning with Diffusion Models}, 
      author={Zhengbang Zhu and Minghuan Liu and Liyuan Mao and Bingyi Kang and Minkai Xu and Yong Yu and Stefano Ermon and Weinan Zhang},
      year={2025},
      eprint={2305.17330},
      archivePrefix={arXiv},
      primaryClass={cs.AI},
      url={https://arxiv.org/abs/2305.17330}, 
}

@misc{deng2024newapproachsolvingsmac,
      title={A New Approach to Solving SMAC Task: Generating Decision Tree Code from Large Language Models}, 
      author={Yue Deng and Weiyu Ma and Yuxin Fan and Yin Zhang and Haifeng Zhang and Jian Zhao},
      year={2024},
      eprint={2410.16024},
      archivePrefix={arXiv},
      primaryClass={cs.AI},
      url={https://arxiv.org/abs/2410.16024}, 
}

@article{Li_Zhao_Wu_Pajarinen_2024, title={Backpropagation Through Agents}, volume={38}, url={https://ojs.aaai.org/index.php/AAAI/article/view/29277}, DOI={10.1609/aaai.v38i12.29277}, abstractNote={A fundamental challenge in multi-agent reinforcement learning (MARL) is to learn the joint policy in an extremely large search space, which grows exponentially with the number of agents. Moreover, fully decentralized policy factorization significantly restricts the search space, which may lead to sub-optimal policies. In contrast, the auto-regressive joint policy can represent a much richer class of joint policies by factorizing the joint policy into the product of a series of conditional individual policies. While such factorization introduces the action dependency among agents explicitly in sequential execution, it does not take full advantage of the dependency during learning. In particular, the subsequent agents do not give the preceding agents feedback about their decisions. In this paper, we propose a new framework Back-Propagation Through Agents (BPTA) that directly accounts for both agents’ own policy updates and the learning of their dependent counterparts. This is achieved by propagating the feedback through action chains. With the proposed framework, our Bidirectional Proximal Policy Optimisation (BPPO) outperforms the state-of-the-art methods. Extensive experiments on matrix games, StarCraftII v2, Multi-agent MuJoCo, and Google Research Football demonstrate the effectiveness of the proposed method.}, number={12}, journal={Proceedings of the AAAI Conference on Artificial Intelligence}, author={Li, Zhiyuan and Zhao, Wenshuai and Wu, Lijun and Pajarinen, Joni}, year={2024}, month={Mar.}, pages={13718-13726} }

@misc{li2024agentmixermultiagentcorrelatedpolicy,
      title={AgentMixer: Multi-Agent Correlated Policy Factorization}, 
      author={Zhiyuan Li and Wenshuai Zhao and Lijun Wu and Joni Pajarinen},
      year={2024},
      eprint={2401.08728},
      archivePrefix={arXiv},
      primaryClass={cs.MA},
      url={https://arxiv.org/abs/2401.08728}, 
}

@misc{mcclellan2024boostingsampleefficiencygeneralization,
      title={Boosting Sample Efficiency and Generalization in Multi-agent Reinforcement Learning via Equivariance}, 
      author={Joshua McClellan and Naveed Haghani and John Winder and Furong Huang and Pratap Tokekar},
      year={2024},
      eprint={2410.02581},
      archivePrefix={arXiv},
      primaryClass={cs.LG},
      url={https://arxiv.org/abs/2410.02581}, 
}

@inproceedings{10.5555/3545946.3598961,
author = {Formanek, Claude and Jeewa, Asad and Shock, Jonathan and Pretorius, Arnu},
title = {Off-the-Grid MARL: Datasets and Baselines for Offline Multi-Agent Reinforcement Learning},
year = {2023},
isbn = {9781450394321},
publisher = {International Foundation for Autonomous Agents and Multiagent Systems},
address = {Richland, SC},
booktitle = {Proceedings of the 2023 International Conference on Autonomous Agents and Multiagent Systems},
pages = {2442–2444},
numpages = {3},
location = {London, United Kingdom},
series = {AAMAS '23}
}

@inproceedings{
wang2021qplex,
title={{\{}QPLEX{\}}: Duplex Dueling Multi-Agent Q-Learning},
author={Jianhao Wang and Zhizhou Ren and Terry Liu and Yang Yu and Chongjie Zhang},
booktitle={International Conference on Learning Representations},
year={2021},
url={https://openreview.net/forum?id=Rcmk0xxIQV}
}

@InProceedings{pmlr-v97-son19a,
  title = 	 {{QTRAN}: Learning to Factorize with Transformation for Cooperative Multi-Agent Reinforcement Learning},
  author =       {Son, Kyunghwan and Kim, Daewoo and Kang, Wan Ju and Hostallero, David Earl and Yi, Yung},
  booktitle = 	 {Proceedings of the 36th International Conference on Machine Learning},
  pages = 	 {5887--5896},
  year = 	 {2019},
  editor = 	 {Chaudhuri, Kamalika and Salakhutdinov, Ruslan},
  volume = 	 {97},
  series = 	 {Proceedings of Machine Learning Research},
  month = 	 {09--15 Jun},
  publisher =    {PMLR},
  url = 	 {https://proceedings.mlr.press/v97/son19a.html}
}

@inproceedings{
lo2024learning,
title={Learning Multi-Agent Communication with Contrastive Learning},
author={Yat Long Lo and Biswa Sengupta and Jakob Nicolaus Foerster and Michael Noukhovitch},
booktitle={The Twelfth International Conference on Learning Representations},
year={2024},
url={https://openreview.net/forum?id=vZZ4hhniJU}
}

@inbook{10.5555/3454287.3454971,
author = {Mahajan, Anuj and Rashid, Tabish and Samvelyan, Mikayel and Whiteson, Shimon},
title = {MAVEN: multi-agent variational exploration},
year = {2019},
publisher = {Curran Associates Inc.},
address = {Red Hook, NY, USA},
booktitle = {Proceedings of the 33rd International Conference on Neural Information Processing Systems},
articleno = {684},
numpages = {12}
}

@inproceedings{10.5555/3600270.3602070,
author = {Wei, Jason and Wang, Xuezhi and Schuurmans, Dale and Bosma, Maarten and Ichter, Brian and Xia, Fei and Chi, Ed H. and Le, Quoc V. and Zhou, Denny},
title = {Chain-of-thought prompting elicits reasoning in large language models},
year = {2024},
isbn = {9781713871088},
publisher = {Curran Associates Inc.},
address = {Red Hook, NY, USA},
booktitle = {Proceedings of the 36th International Conference on Neural Information Processing Systems},
articleno = {1800},
numpages = {14},
location = {New Orleans, LA, USA},
series = {NIPS '22}
}

@article{Besta_Blach_Kubicek_Gerstenberger_Podstawski_Gianinazzi_Gajda_Lehmann_Niewiadomski_Nyczyk_Hoefler_2024, title={Graph of Thoughts: Solving Elaborate Problems with Large Language Models}, volume={38}, url={https://ojs.aaai.org/index.php/AAAI/article/view/29720}, DOI={10.1609/aaai.v38i16.29720}, abstractNote={We introduce Graph of Thoughts (GoT): a framework that
advances prompting capabilities in large language models
(LLMs) beyond those offered by paradigms such as Chain-of-Thought or Tree of Thoughts (ToT). The key idea and primary advantage of GoT is the ability to model the information generated by an LLM as an arbitrary graph, where units of information (&quot;LLM thoughts&quot;) are vertices, and edges correspond
to dependencies between these vertices. This approach enables combining arbitrary LLM thoughts into synergistic outcomes, distilling the essence of whole networks of thoughts,
or enhancing thoughts using feedback loops. We illustrate
that GoT offers advantages over state of the art on different
tasks, for example increasing the quality of sorting by 62%
over ToT, while simultaneously reducing costs by &gt;31%.
We ensure that GoT is extensible with new thought transformations and thus can be used to spearhead new prompting
schemes. This work brings the LLM reasoning closer to human thinking or brain mechanisms such as recurrence, both
of which form complex networks}, number={16}, journal={Proceedings of the AAAI Conference on Artificial Intelligence}, author={Besta, Maciej and Blach, Nils and Kubicek, Ales and Gerstenberger, Robert and Podstawski, Michal and Gianinazzi, Lukas and Gajda, Joanna and Lehmann, Tomasz and Niewiadomski, Hubert and Nyczyk, Piotr and Hoefler, Torsten}, year={2024}, month={Mar.}, pages={17682-17690} }

@inproceedings{10.5555/3600270.3601883,
author = {Kojima, Takeshi and Gu, Shixiang Shane and Reid, Machel and Matsuo, Yutaka and Iwasawa, Yusuke},
title = {Large language models are zero-shot reasoners},
year = {2024},
isbn = {9781713871088},
publisher = {Curran Associates Inc.},
address = {Red Hook, NY, USA},
booktitle = {Proceedings of the 36th International Conference on Neural Information Processing Systems},
articleno = {1613},
numpages = {15},
location = {New Orleans, LA, USA},
series = {NIPS '22}
}

@inproceedings{10.5555/3666122.3667509,
author = {Zhao, Zirui and Lee, Wee Sun and Hsu, David},
title = {Large language models as commonsense knowledge for large-scale task planning},
year = {2024},
publisher = {Curran Associates Inc.},
address = {Red Hook, NY, USA},
booktitle = {Proceedings of the 37th International Conference on Neural Information Processing Systems},
articleno = {1387},
numpages = {21},
location = {New Orleans, LA, USA},
series = {NIPS '23}
}

@article{
wang2024voyager,
title={Voyager: An Open-Ended Embodied Agent with Large Language Models},
author={Guanzhi Wang and Yuqi Xie and Yunfan Jiang and Ajay Mandlekar and Chaowei Xiao and Yuke Zhu and Linxi Fan and Anima Anandkumar},
journal={Transactions on Machine Learning Research},
issn={2835-8856},
year={2024},
url={https://openreview.net/forum?id=ehfRiF0R3a},
note={}
}

@inproceedings{
yao2023react,
title={ReAct: Synergizing Reasoning and Acting in Language Models},
author={Shunyu Yao and Jeffrey Zhao and Dian Yu and Nan Du and Izhak Shafran and Karthik R Narasimhan and Yuan Cao},
booktitle={The Eleventh International Conference on Learning Representations },
year={2023},
url={https://openreview.net/forum?id=WE_vluYUL-X}
}

@inproceedings{NEURIPS2023_1b44b878,
 author = {Shinn, Noah and Cassano, Federico and Gopinath, Ashwin and Narasimhan, Karthik and Yao, Shunyu},
 booktitle = {Advances in Neural Information Processing Systems},
 editor = {A. Oh and T. Naumann and A. Globerson and K. Saenko and M. Hardt and S. Levine},
 pages = {8634--8652},
 publisher = {Curran Associates, Inc.},
 title = {Reflexion: language agents with verbal reinforcement learning},
 url = {https://proceedings.neurips.cc/paper_files/paper/2023/file/1b44b878bb782e6954cd888628510e90-Paper-Conference.pdf},
 volume = {36},
 year = {2023}
}

@inproceedings{
ichter2022do,
title={Do As I Can, Not As I Say: Grounding Language in Robotic Affordances},
author={brian ichter and Anthony Brohan and Yevgen Chebotar and Chelsea Finn and Karol Hausman and Alexander Herzog and Daniel Ho and Julian Ibarz and Alex Irpan and Eric Jang and Ryan Julian and Dmitry Kalashnikov and Sergey Levine and Yao Lu and Carolina Parada and Kanishka Rao and Pierre Sermanet and Alexander T Toshev and Vincent Vanhoucke and Fei Xia and Ted Xiao and Peng Xu and Mengyuan Yan and Noah Brown and Michael Ahn and Omar Cortes and Nicolas Sievers and Clayton Tan and Sichun Xu and Diego Reyes and Jarek Rettinghouse and Jornell Quiambao and Peter Pastor and Linda Luu and Kuang-Huei Lee and Yuheng Kuang and Sally Jesmonth and Kyle Jeffrey and Rosario Jauregui Ruano and Jasmine Hsu and Keerthana Gopalakrishnan and Byron David and Andy Zeng and Chuyuan Kelly Fu},
booktitle={6th Annual Conference on Robot Learning},
year={2022},
url={https://openreview.net/forum?id=bdHkMjBJG_w}
}

@misc{ma2024largelanguagemodelsplay,
      title={Large Language Models Play StarCraft II: Benchmarks and A Chain of Summarization Approach}, 
      author={Weiyu Ma and Qirui Mi and Yongcheng Zeng and Xue Yan and Yuqiao Wu and Runji Lin and Haifeng Zhang and Jun Wang},
      year={2024},
      eprint={2312.11865},
      archivePrefix={arXiv},
      primaryClass={cs.AI},
      url={https://arxiv.org/abs/2312.11865}, 
}

@misc{wang2024executablecodeactionselicit,
      title={Executable Code Actions Elicit Better LLM Agents}, 
      author={Xingyao Wang and Yangyi Chen and Lifan Yuan and Yizhe Zhang and Yunzhu Li and Hao Peng and Heng Ji},
      year={2024},
      eprint={2402.01030},
      archivePrefix={arXiv},
      primaryClass={cs.CL},
      url={https://arxiv.org/abs/2402.01030}, 
}

@inproceedings{10.5555/2283396.2283451,
author = {Pajarinen, Joni and Peltonen, Jaakko},
title = {Efficient planning for factored infinite-horizon DEC-POMDPs},
year = {2011},
isbn = {9781577355137},
publisher = {AAAI Press},
booktitle = {Proceedings of the Twenty-Second International Joint Conference on Artificial Intelligence - Volume Volume One},
pages = {325–331},
numpages = {7},
location = {Barcelona, Catalonia, Spain},
series = {IJCAI'11}
}

@misc{nayak2025llamarlonghorizonplanningmultiagent,
      title={LLaMAR: Long-Horizon Planning for Multi-Agent Robots in Partially Observable Environments}, 
      author={Siddharth Nayak and Adelmo Morrison Orozco and Marina Ten Have and Vittal Thirumalai and Jackson Zhang and Darren Chen and Aditya Kapoor and Eric Robinson and Karthik Gopalakrishnan and James Harrison and Brian Ichter and Anuj Mahajan and Hamsa Balakrishnan},
      year={2025},
      eprint={2407.10031},
      archivePrefix={arXiv},
      primaryClass={cs.RO},
      url={https://arxiv.org/abs/2407.10031}, 
}

@INPROCEEDINGS{10610855,
  author={Mandi, Zhao and Jain, Shreeya and Song, Shuran},
  booktitle={2024 IEEE International Conference on Robotics and Automation (ICRA)}, 
  title={RoCo: Dialectic Multi-Robot Collaboration with Large Language Models}, 
  year={2024},
  volume={},
  number={},
  pages={286-299},
  doi={10.1109/ICRA57147.2024.10610855}}

@misc{zhang2024buildingcooperativeembodiedagents,
      title={Building Cooperative Embodied Agents Modularly with Large Language Models}, 
      author={Hongxin Zhang and Weihua Du and Jiaming Shan and Qinhong Zhou and Yilun Du and Joshua B. Tenenbaum and Tianmin Shu and Chuang Gan},
      year={2024},
      eprint={2307.02485},
      archivePrefix={arXiv},
      primaryClass={cs.AI},
      url={https://arxiv.org/abs/2307.02485}, 
}

@inproceedings{gong-etal-2024-mindagent,
    title = "{M}ind{A}gent: Emergent Gaming Interaction",
    author = "Gong, Ran  and
      Huang, Qiuyuan  and
      Ma, Xiaojian  and
      Noda, Yusuke  and
      Durante, Zane  and
      Zheng, Zilong  and
      Terzopoulos, Demetri  and
      Fei-Fei, Li  and
      Gao, Jianfeng  and
      Vo, Hoi",
    editor = "Duh, Kevin  and
      Gomez, Helena  and
      Bethard, Steven",
    booktitle = "Findings of the Association for Computational Linguistics: NAACL 2024",
    month = jun,
    year = "2024",
    address = "Mexico City, Mexico",
    publisher = "Association for Computational Linguistics",
    url = "https://aclanthology.org/2024.findings-naacl.200/",
    doi = "10.18653/v1/2024.findings-naacl.200",
    pages = "3154--3183"
}

@InProceedings{Puig_2018_CVPR,
author = {Puig, Xavier and Ra, Kevin and Boben, Marko and Li, Jiaman and Wang, Tingwu and Fidler, Sanja and Torralba, Antonio},
title = {VirtualHome: Simulating Household Activities via Programs},
booktitle = {Proceedings of the IEEE Conference on Computer Vision and Pattern Recognition (CVPR)},
month = {June},
year = {2018}
}

@misc{tan2024cradleempoweringfoundationagents,
      title={Cradle: Empowering Foundation Agents Towards General Computer Control}, 
      author={Weihao Tan and Wentao Zhang and Xinrun Xu and Haochong Xia and Ziluo Ding and Boyu Li and Bohan Zhou and Junpeng Yue and Jiechuan Jiang and Yewen Li and Ruyi An and Molei Qin and Chuqiao Zong and Longtao Zheng and Yujie Wu and Xiaoqiang Chai and Yifei Bi and Tianbao Xie and Pengjie Gu and Xiyun Li and Ceyao Zhang and Long Tian and Chaojie Wang and Xinrun Wang and Börje F. Karlsson and Bo An and Shuicheng Yan and Zongqing Lu},
      year={2024},
      eprint={2403.03186},
      archivePrefix={arXiv},
      primaryClass={cs.AI},
      url={https://arxiv.org/abs/2403.03186}, 
}

@inproceedings{10.1609/aaai.v38i16.29710,
author = {Zhang, Ceyao and Yang, Kaijie and Hu, Siyi and Wang, Zihao and Li, Guanghe and Sun, Yihang and Zhang, Cheng and Zhang, Zhaowei and Liu, Anji and Zhu, Song-Chun and Chang, Xiaojun and Zhang, Junge and Yin, Feng and Liang, Yitao and Yang, Yaodong},
title = {ProAgent: building proactive cooperative agents with large language models},
year = {2025},
isbn = {978-1-57735-887-9},
publisher = {AAAI Press},
url = {https://doi.org/10.1609/aaai.v38i16.29710},
doi = {10.1609/aaai.v38i16.29710},
booktitle = {Proceedings of the Thirty-Eighth AAAI Conference on Artificial Intelligence and Thirty-Sixth Conference on Innovative Applications of Artificial Intelligence and Fourteenth Symposium on Educational Advances in Artificial Intelligence},
articleno = {1962},
numpages = {9},
series = {AAAI'24/IAAI'24/EAAI'24}
}

@misc{gao2024coohoilearningcooperativehumanobject,
      title={CooHOI: Learning Cooperative Human-Object Interaction with Manipulated Object Dynamics}, 
      author={Jiawei Gao and Ziqin Wang and Zeqi Xiao and Jingbo Wang and Tai Wang and Jinkun Cao and Xiaolin Hu and Si Liu and Jifeng Dai and Jiangmiao Pang},
      year={2024},
      eprint={2406.14558},
      archivePrefix={arXiv},
      primaryClass={cs.RO},
      url={https://arxiv.org/abs/2406.14558}, 
}

@misc{feng2024learningmultiagentlocomanipulationlonghorizon,
      title={Learning Multi-Agent Loco-Manipulation for Long-Horizon Quadrupedal Pushing}, 
      author={Yuming Feng and Chuye Hong and Yaru Niu and Shiqi Liu and Yuxiang Yang and Wenhao Yu and Tingnan Zhang and Jie Tan and Ding Zhao},
      year={2024},
      eprint={2411.07104},
      archivePrefix={arXiv},
      primaryClass={cs.RO},
      url={https://arxiv.org/abs/2411.07104}, 
}

@inproceedings{
monroc2024wfcrl,
title={{WFCRL}: A Multi-Agent Reinforcement Learning Benchmark for Wind Farm Control},
author={Claire Bizon Monroc and Ana Busic and Donatien Dubuc and Jiamin Zhu},
booktitle={The Thirty-eight Conference on Neural Information Processing Systems Datasets and Benchmarks Track},
year={2024},
url={https://openreview.net/forum?id=ZRMAhpZ3ED}
}

@article{Kurach_Raichuk_Stańczyk_Zając_Bachem_Espeholt_Riquelme_Vincent_Michalski_Bousquet_Gelly_2020, title={Google Research Football: A Novel Reinforcement Learning Environment}, volume={34}, url={https://ojs.aaai.org/index.php/AAAI/article/view/5878}, DOI={10.1609/aaai.v34i04.5878}, abstractNote={&lt;p&gt;Recent progress in the field of reinforcement learning has been accelerated by virtual learning environments such as video games, where novel algorithms and ideas can be quickly tested in a safe and reproducible manner. We introduce the &lt;em&gt;Google Research Football Environment&lt;/em&gt;, a new reinforcement learning environment where agents are trained to play football in an advanced, physics-based 3D simulator. The resulting environment is challenging, easy to use and customize, and it is available under a permissive open-source license. In addition, it provides support for multiplayer and multi-agent experiments. We propose three full-game scenarios of varying difficulty with the &lt;em&gt;Football Benchmarks&lt;/em&gt; and report baseline results for three commonly used reinforcement algorithms (IMPALA, PPO, and Ape-X DQN). We also provide a diverse set of simpler scenarios with the &lt;em&gt;Football Academy&lt;/em&gt; and showcase several promising research directions.&lt;/p&gt;}, number={04}, journal={Proceedings of the AAAI Conference on Artificial Intelligence}, author={Kurach, Karol and Raichuk, Anton and Stanczyk, Piotr and Zajac, Michał and Bachem, Olivier and Espeholt, Lasse and Riquelme, Carlos and Vincent, Damien and Michalski, Marcin and Bousquet, Olivier and Gelly, Sylvain}, year={2020}, month={Apr.}, pages={4501-4510} }

@inproceedings{10.5555/3295222.3295385,
author = {Lowe, Ryan and Wu, Yi and Tamar, Aviv and Harb, Jean and Abbeel, Pieter and Mordatch, Igor},
title = {Multi-agent actor-critic for mixed cooperative-competitive environments},
year = {2017},
isbn = {9781510860964},
publisher = {Curran Associates Inc.},
address = {Red Hook, NY, USA},
booktitle = {Proceedings of the 31st International Conference on Neural Information Processing Systems},
pages = {6382–6393},
numpages = {12},
location = {Long Beach, California, USA},
series = {NIPS'17}
}

@article{
su2024a,
title={A Fully Decentralized Surrogate for Multi-Agent Policy Optimization},
author={Kefan Su and Zongqing Lu},
journal={Transactions on Machine Learning Research},
issn={2835-8856},
year={2024},
url={https://openreview.net/forum?id=MppUW90uU2},
note={}
}

@InProceedings{pmlr-v80-zhang18n,
  title = 	 {Fully Decentralized Multi-Agent Reinforcement Learning with Networked Agents},
  author =       {Zhang, Kaiqing and Yang, Zhuoran and Liu, Han and Zhang, Tong and Basar, Tamer},
  booktitle = 	 {Proceedings of the 35th International Conference on Machine Learning},
  pages = 	 {5872--5881},
  year = 	 {2018},
  editor = 	 {Dy, Jennifer and Krause, Andreas},
  volume = 	 {80},
  series = 	 {Proceedings of Machine Learning Research},
  month = 	 {10--15 Jul},
  publisher =    {PMLR},
  url = 	 {https://proceedings.mlr.press/v80/zhang18n.html}
}

@article{ma2024efficient,
  title={Efficient and scalable reinforcement learning for large-scale network control},
  author={Ma, Chengdong and Li, Aming and Du, Yali and Dong, Hao and Yang, Yaodong},
  journal={Nature Machine Intelligence},
  volume={6},
  number={9},
  pages={1006--1020},
  year={2024},
  publisher={Nature Publishing Group UK London}
}

@inbook{10.5555/3454287.3454752,
author = {Carroll, Micah and Shah, Rohin and Ho, Mark K. and Griffiths, Thomas L. and Seshia, Sanjit A. and Abbeel, Pieter and Dragan, Anca},
title = {On the utility of learning about humans for human-AI coordination},
year = {2019},
publisher = {Curran Associates Inc.},
address = {Red Hook, NY, USA},
booktitle = {Proceedings of the 33rd International Conference on Neural Information Processing Systems},
articleno = {465},
numpages = {12}
}

@inproceedings{10.5555/3666122.3668386,
author = {Li, Guohao and Al Kader Hammoud, Hasan Abed and Itani, Hani and Khizbullin, Dmitrii and Ghanem, Bernard},
title = {CAMEL: communicative agents for "mind" exploration of large language model society},
year = {2024},
publisher = {Curran Associates Inc.},
address = {Red Hook, NY, USA},
booktitle = {Proceedings of the 37th International Conference on Neural Information Processing Systems},
articleno = {2264},
numpages = {18},
location = {New Orleans, LA, USA},
series = {NIPS '23}
}

@InProceedings{pmlr-v139-iqbal21a,
  title = 	 {Randomized Entity-wise Factorization for Multi-Agent Reinforcement Learning},
  author =       {Iqbal, Shariq and De Witt, Christian A Schroeder and Peng, Bei and Boehmer, Wendelin and Whiteson, Shimon and Sha, Fei},
  booktitle = 	 {Proceedings of the 38th International Conference on Machine Learning},
  pages = 	 {4596--4606},
  year = 	 {2021},
  editor = 	 {Meila, Marina and Zhang, Tong},
  volume = 	 {139},
  series = 	 {Proceedings of Machine Learning Research},
  month = 	 {18--24 Jul},
  publisher =    {PMLR},
  url = 	 {https://proceedings.mlr.press/v139/iqbal21a.html}
}

@InProceedings{pmlr-v202-ding23d,
  title = 	 {Entity Divider with Language Grounding in Multi-Agent Reinforcement Learning},
  author =       {Ding, Ziluo and Zhang, Wanpeng and Yue, Junpeng and Wang, Xiangjun and Huang, Tiejun and Lu, Zongqing},
  booktitle = 	 {Proceedings of the 40th International Conference on Machine Learning},
  pages = 	 {8103--8119},
  year = 	 {2023},
  editor = 	 {Krause, Andreas and Brunskill, Emma and Cho, Kyunghyun and Engelhardt, Barbara and Sabato, Sivan and Scarlett, Jonathan},
  volume = 	 {202},
  series = 	 {Proceedings of Machine Learning Research},
  month = 	 {23--29 Jul},
  publisher =    {PMLR},
  url = 	 {https://proceedings.mlr.press/v202/ding23d.html}
}

@misc{wang2024qwen2vlenhancingvisionlanguagemodels,
      title={Qwen2-VL: Enhancing Vision-Language Model's Perception of the World at Any Resolution}, 
      author={Peng Wang and Shuai Bai and Sinan Tan and Shijie Wang and Zhihao Fan and Jinze Bai and Keqin Chen and Xuejing Liu and Jialin Wang and Wenbin Ge and Yang Fan and Kai Dang and Mengfei Du and Xuancheng Ren and Rui Men and Dayiheng Liu and Chang Zhou and Jingren Zhou and Junyang Lin},
      year={2024},
      eprint={2409.12191},
      archivePrefix={arXiv},
      primaryClass={cs.CV},
      url={https://arxiv.org/abs/2409.12191}, 
}

@inproceedings{10.1145/3613905.3651029,
author = {Wang, Chao and Hasler, Stephan and Tanneberg, Daniel and Ocker, Felix and Joublin, Frank and Ceravola, Antonello and Deigmoeller, Joerg and Gienger, Michael},
title = {LaMI: Large Language Models for Multi-Modal Human-Robot Interaction},
year = {2024},
isbn = {9798400703317},
publisher = {Association for Computing Machinery},
address = {New York, NY, USA},
url = {https://doi.org/10.1145/3613905.3651029},
doi = {10.1145/3613905.3651029},
booktitle = {Extended Abstracts of the CHI Conference on Human Factors in Computing Systems},
articleno = {218},
numpages = {10},
location = {Honolulu, HI, USA},
series = {CHI EA '24}
}

@misc{liang2023codepolicieslanguagemodel,
      title={Code as Policies: Language Model Programs for Embodied Control}, 
      author={Jacky Liang and Wenlong Huang and Fei Xia and Peng Xu and Karol Hausman and Brian Ichter and Pete Florence and Andy Zeng},
      year={2023},
      eprint={2209.07753},
      archivePrefix={arXiv},
      primaryClass={cs.RO},
      url={https://arxiv.org/abs/2209.07753}, 
}

@inproceedings{
hong2024metagpt,
title={Meta{GPT}: Meta Programming for A Multi-Agent Collaborative Framework},
author={Sirui Hong and Mingchen Zhuge and Jonathan Chen and Xiawu Zheng and Yuheng Cheng and Jinlin Wang and Ceyao Zhang and Zili Wang and Steven Ka Shing Yau and Zijuan Lin and Liyang Zhou and Chenyu Ran and Lingfeng Xiao and Chenglin Wu and J{\"u}rgen Schmidhuber},
booktitle={The Twelfth International Conference on Learning Representations},
year={2024},
url={https://openreview.net/forum?id=VtmBAGCN7o}
}

@inproceedings{formanekoryx,
  title={Oryx: a Scalable Sequence Model for Many-Agent Coordination in Offline MARL},
  author={Formanek, Juan Claude and Mahjoub, Omayma and Nessir, Louay Ben and Abramowitz, Sasha and de Kock, Ruan John and Khlifi, Wiem and Rajaonarivonivelomanantsoa, Daniel and Du Toit, Simon Verster and Fokam, Arnol Manuel and Singh, Siddarth and others},
  booktitle={The Thirty-ninth Annual Conference on Neural Information Processing Systems}
}


\newpage
\appendix
\onecolumn
\section{Pseudocode}
The pseudo-code of the COMPASS algorithm is shown in Pseudocode 1.
\begin{algorithm}[htbp]
\caption{COMPASS Agent Decision-Making Loop}
\label{alg:compass}
\begin{algorithmic}
\STATE \textbf{Initialize:}
\STATE skill\_manager.bootstrap(demonstration\_data)
\STATE agent\_state $\gets$ environment.reset(agent\_id)
\STATE local\_memory $\gets$ initialize\_local\_memory()
\STATE global\_memory $\gets$ initialize\_global\_memory()
\STATE previous\_action\_result $\gets$ None
\WHILE{True}
    \STATE \COMMENT{1. Communication Phase}
    \STATE local\_observations $\gets$ agent\_state.get\_observations()
    \STATE communication\_protocol.share\_local\_observations(agent\_id, local\_observations, global\_memory)
    \STATE global\_entity\_info $\gets$ communication\_protocol.get\_global\_memory.update(global\_memory)
    
    \STATE \COMMENT{2. Perception Phase}
    \STATE processed\_state $\gets$ vlm\_perception.process(raw\_observation=agent\_state,
    \STATE \hspace{2em} communication\_data=global\_entity\_info, local\_memory=local\_memory)
    \STATE local\_memory.update(processed\_state)
    
    \STATE \COMMENT{3. Self-Reflection Phase}
    \IF{previous\_action\_result $\neq$ None}
        \STATE reflection\_feedback $\gets$ vlm\_self\_reflection.reflect(previous\_action\_result)
    \ENDIF
    
    \STATE \COMMENT{4. Task Reasoning Phase}
    \STATE sub\_task $\gets$ vlm\_task\_reasoning.props\_subtask(processed\_state, overall\_goal, reflection\_feedback)
    
    \STATE \COMMENT{5. Skill Generation Phase}
    \STATE new\_skill\_code $\gets$ vlm\_skill\_generator.generate\_skill(sub\_task, processed\_state)
    \STATE skill\_manager.add\_skill(new\_skill\_code, sub\_task)
    
    \STATE \COMMENT{6. Actor Phase}
    \STATE relevant\_skills $\gets$ skill\_manager.retrieve\_skills(sub\_task)
    \STATE chosen\_skill\_code $\gets$ vlm\_actor.select\_skill(sub\_task, relevant\_skills, processed\_state)
    
    \STATE \COMMENT{7. Execution Phase}
    \STATE (next\_agent\_state, reward, done, info) $\gets$ environment.step(agent\_id, chosen\_skill\_code)
    \STATE agent\_state $\gets$ next\_agent\_state
    \STATE previous\_action\_result $\gets$ (chosen\_skill\_code, info, reward, done)
    \STATE local\_memory.add\_action(chosen\_skill\_code)
    
    \IF{done}
        \STATE \textbf{break}
    \ENDIF
\ENDWHILE
\end{algorithmic}
\end{algorithm}

\section{Implementation Details}
COMPASS integrates VLMs to process multi-modal inputs and generate executable skills in two stages. Each skill follows a standardized interface:
\begin{lstlisting}[caption={Interface of Generated skills.}, label={lst:example}]
def skill_template(obs: str):  
    obs_data = parse_obs(obs) 
    def score_target(unit):
        ...
        return score
    def control_logic():
        ...
        return atomic_action
\end{lstlisting}
The skills follow a structured decision-making pipeline with two core components: score\_target(unit) and control\_logic():
\begin{itemize}
    \item score\_target(unit): Dynamically calculates a threat/priority score by evaluating unit type, health, distance, formations, and matchups to guide optimal attack/heal targeting decisions.
    \item control\_logic(): Dynamically coordinates unit behavior by integrating observations, target priorities, and pathfinding to execute role-optimized strategies (e.g., stalkers attack colossus at max range while moving away from zealots).
\end{itemize}

COMPASS evolves skills through iterative refinement and task-guided synthesis:
\begin{itemize}
    \item iterative refinement: When errors occur during skill execution, VLMs analyze the error messages and attempt to fix the bugs.
    \item task-guided synthesis: When a new task is proposed, VLMs first determine whether a new skill needs to be generated to align with the task. If necessary, VLMs generate new score\_target or control\_logic components to fulfill the task requirements and integrate them with the existing code to construct a new skill.
\end{itemize}

For example, if the task is: "\textit{Implement an aggressive advance movement pattern for the colossus unit when enemy stalkers are within sight range and allies are positioned to provide covering fire}," and the current skills in the library are not aggressive enough, the VLMs will refine the control\_logic to implement a more aggressive behavior pattern.

COMPASS maintains a global entity memory, where agents act as nodes connected if within sight range. Each node stores information about visible allies and enemies. When Agent A observes entities, it propagates updates through adjacent nodes recursively, up to max\_hops (default = 3). This allows agents to infer off-screen threats via ally intermediaries. For implementation details, see ./common/memory/global\_memory.py.
\section{Environment Settings and More Results}

\begin{table}[ht]
\caption{We report quantitative results on SMACv2 under sparse reward settings, excluding COMPASS due to its inherent insensitivity to reward sparsity.}
\centering
\begin{tabular}{lcccc}
\hline
& QMIX$_{s}$ & MAPPO$_{s}$ & HAPPO$_{s}$ & HASAC$_{s}$ \\
\hline
\multicolumn{5}{c}{PROTOSS} \\
\hline
5V5 & 0 & 0 & 0 & 0 \\
5V6 & 0 & 0 & 0 & 0 \\
\hline
\multicolumn{5}{c}{TERRAN} \\
\hline
5V5 & 0 & 0 & 0 & 0 \\
5V6 & 0 & 0 & 0 & 0 \\
\hline
\multicolumn{5}{c}{ZERG} \\
\hline
5V5 & $0.02_{0.01}$ & 0 & 0 & 0 \\
5V6 & 0 & 0 & 0 & 0 \\
\hline
\end{tabular}
\label{tab:quantitative_results_sparse}
\end{table}

\section{More Skill Analysis}

\begin{figure}
    \begin{center}
        \includegraphics[width=0.99\columnwidth]{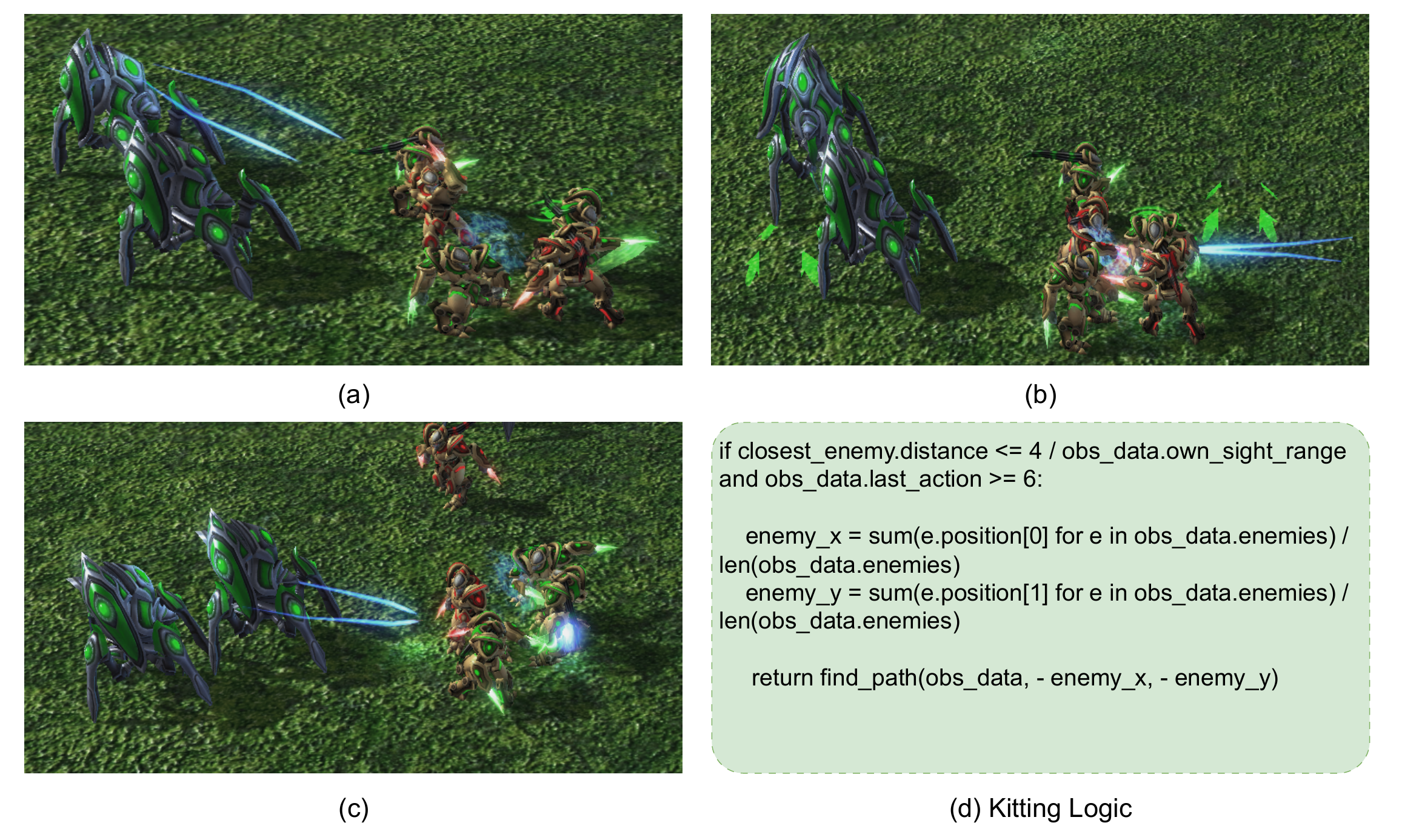}
        \caption{Illustration and implementation of Kitting logic. (a)-(c) demonstrate progressive stages of the kitting tactic where allied units strategically maintain optimal attack range while retreating from melee enemies. (d) shows the corresponding Python code snippet generated by the VLMs.} 
        \Description{Illustration and implementation of Kitting logic. (a)-(c) demonstrate progressive stages of the kitting tactic where allied units strategically maintain optimal attack range while retreating from melee enemies. (d) shows the corresponding Python code snippet generated by the VLMs.}
        \label{fig:kitting}
    \end{center}
\end{figure}

\begin{figure}
    \begin{center}
        \includegraphics[width=0.99\columnwidth]{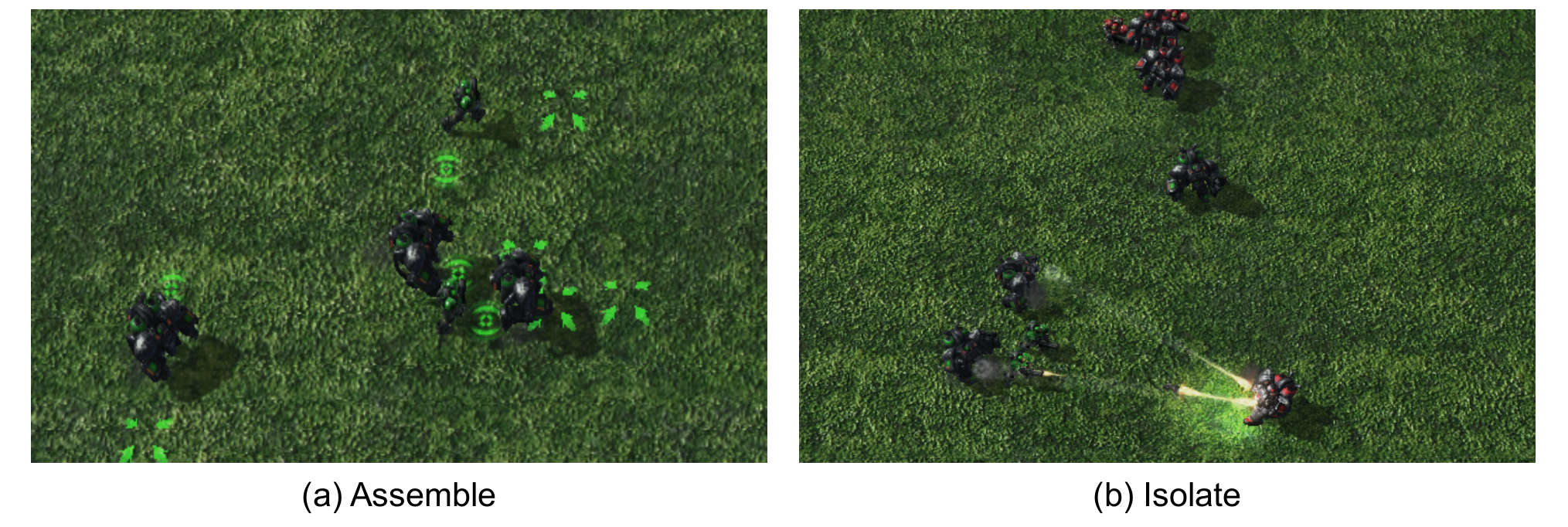}
        \caption{Illustration of Isolating logic. (a) Allied units strategically assemble into a cohesive formation. (b) The assembled units execute a rapid engagement against an isolated enemy unit, eliminating it before reinforcements can arrive, thus creating a numerical advantage.} 
        \Description{Illustration of Isolating logic. (a) Allied units strategically assemble into a cohesive formation. (b) The assembled units execute a rapid engagement against an isolated enemy unit, eliminating it before reinforcements can arrive, thus creating a numerical advantage.}
        \label{fig:isolate}
    \end{center}
\end{figure}
\begin{figure}
    \begin{center}
        \includegraphics[width=0.99\columnwidth]{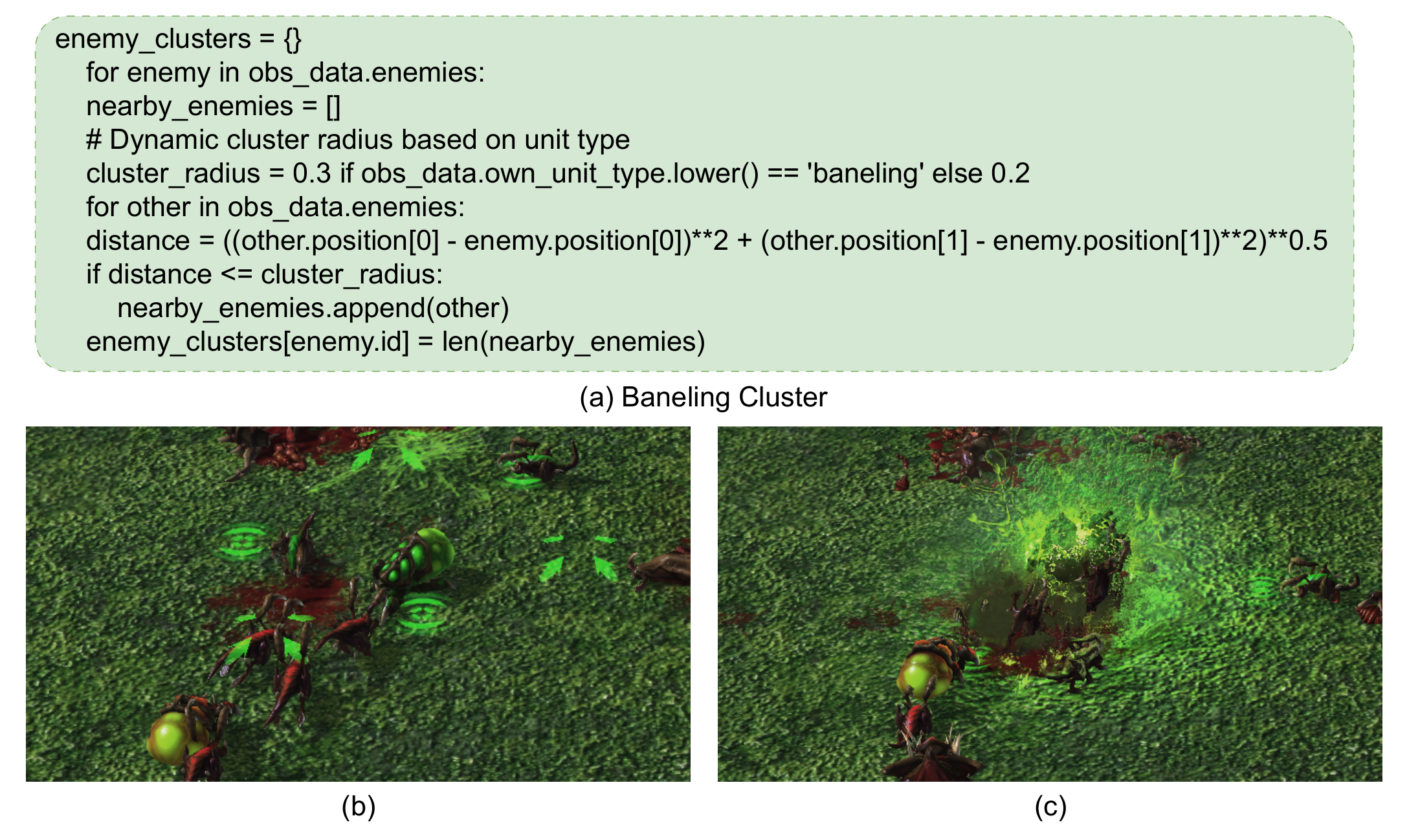}
        \caption{Demonstration of area-of-effect (AOE) optimization for Baneling units in SMACv2. (a) The VLM-generated Python code calculates optimal detonation positions by analyzing enemy cluster density and positions. (b-c) Visual sequence showing Baneling execution, where the unit identifies a dense cluster of enemy units and detonates for maximum AOE damage.} 
        \Description{Demonstration of area-of-effect (AOE) optimization for Baneling units in SMACv2. (a) The VLM-generated Python code calculates optimal detonation positions by analyzing enemy cluster density and positions. (b-c) Visual sequence showing Baneling execution, where the unit identifies a dense cluster of enemy units and detonates for maximum AOE damage.} 
        \label{fig:baneling}
    \end{center}
\end{figure}

\section{Token Usage Analysis}

We acknowledge the computational cost. We provide a breakdown of token usage per step per agent (call frequency: 20).

\begin{itemize}
    \item Average Episode Total Tokens: 80,000
    \item Average Tokens per Step: 1,333 (Perception: 2k, Actor: 2k, Self-Reflection: 2k, Task Reasoning: 10k, Skill Synthesis: 10k)
\end{itemize}

The bottlenecks are Task Reasoning and Skill Synthesis, which require historical context and code history. Skill synthesis often requires multiple rounds of self-correction to generate executable code. Future work will focus on reducing this cost by exploiting structural symmetries among agents. We may also reduce token usage by appropriately lowering the frequency of queries.

\section{Baseline training results}
For QMIX, we utilized Epymarl\footnote{https://github.com/uoe-agents/epymarl}, and for others, we used HARL\footnote{https://github.com/PKU-MARL/HARL}. Task difficulty was set to 5v5 for SYMMETRIC tasks and 5v6 for ASYMMETRIC tasks; the 5v6 scenario introduces a significant force disadvantage (20\% outnumbered vs 10\% in 10v11).
\begin{figure*}[htbp!]
\centering
\includegraphics[width=.95\textwidth]{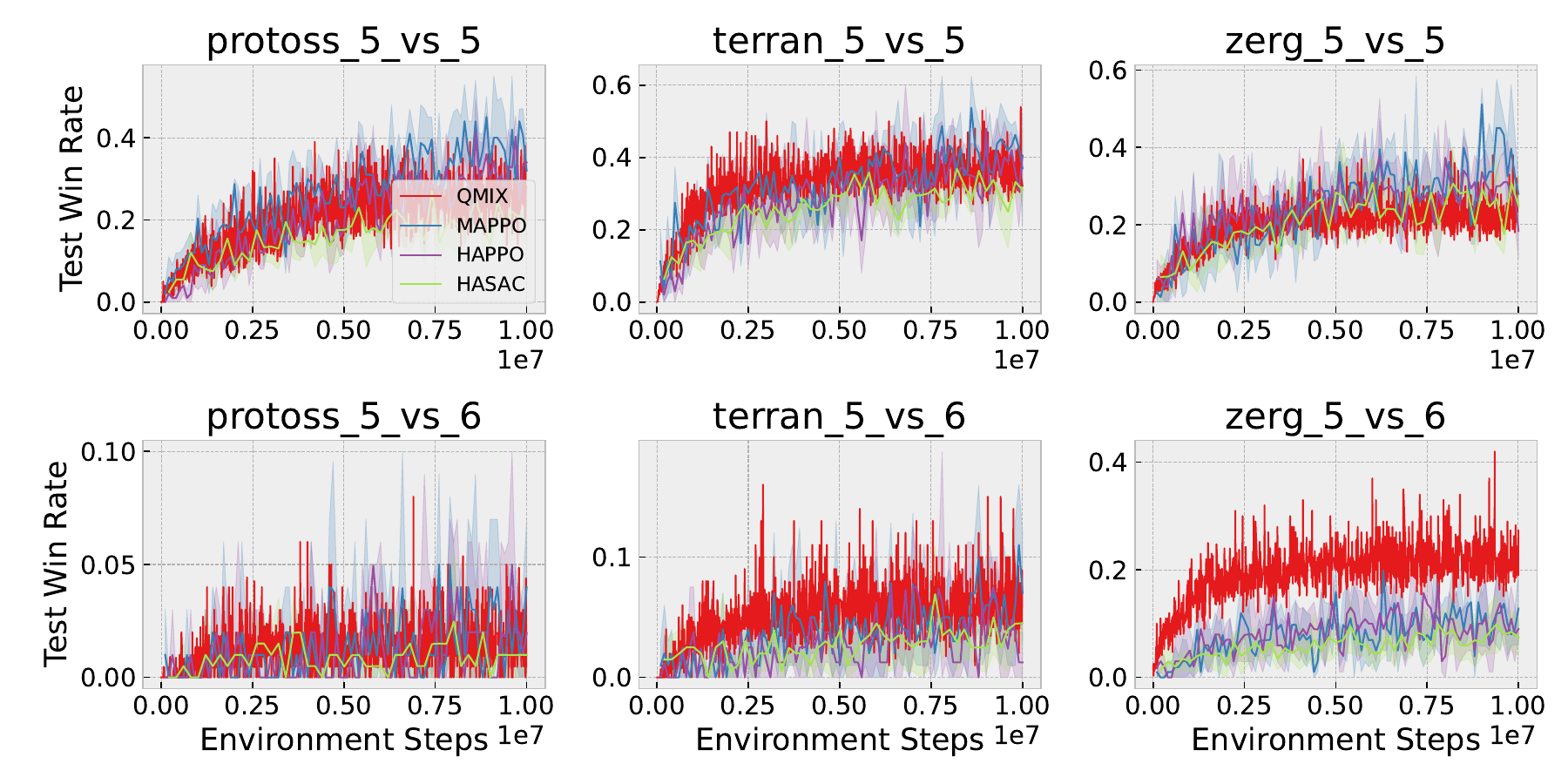} 
\caption{Baseline training results on SMACv2.}
\Description{Baseline training results on SMACv2.}
\label{fig:baselines_results}
\end{figure*}


\newpage
\section{Prompts used in COMPASS}
\begin{tcolorbox}[breakable,colback=black!5,colframe=black!40!black,title=Prompt for Perception]
\small{
You are an AI assistant helping with academic research in the StarCraft II's SMAC (StarCraft Multi-Agent Challenge) environment, controlling a $<$unit\_type$>$ unit with ID $<$unit\_id$>$ in micromanagement scenarios $<$scenario\_name$>$ to help your team defeat the enemy forces. You operate under decentralized execution with partial observability, making decisions based only on local observations within your unit's field of view. Your advanced capabilities enable you to process and interpret gameplay screenshots and other relevant information.\\
I will give you the following information:\\
$<$few\_shots$>$\\
Reasoning for the last episode:\\
$<$last\_episode\_reasoning$>$\\
Strategic situation analysis:\\
$<$info\_summary$>$\\
Below is the current in-game screenshot and its description:\\
$<$image\_introduction$>$\\
Minimap information:\\
$<$ego\_minimap$>$\\
Current task:\\
$<$task\_description$>$\\
Tactics recommendation:\\
$<$web\_search$>$\\
Based on the above information, you should first analyze the current game situation by integrating the information from the in-game screenshot, its description, and other provided information.\\
Game situation:\\
You should think step by step and provide detailed reasoning to determine the current state of the game. You need to answer the following questions step by step:\\
   1. What is your unit\_id, unit\ type?\\
   2. What map borders are you near? Check which cardinal directions (N/S/E/W) have unavailable movement actions.\\
   3. What is the current health status of your unit? What is the current shield status of your unit?\\
   4. Are there any enemy units visible, either in observation or minimap?\\
   5. Are there any ally units visible, either observation or minimap?\\
   6. Are you positioned at the optimal attack range from enemies, or do you need to reposition based on the enemies' locations and directions?\\
Region of interest: \\
What unit or location should be interacted with to complete the task based on the current screenshot and the current task? You should obey the following rules:  \\
   1. If your chosen region of interest is a unit, format the output as "[Enemy/Ally] \#[target\_id]" (e.g., "Enemy \#0" for enemy unit with ID 0, "Ally \#1" for ally unit with ID 1)\\
   2. If your chosen region of interest is location, format the output as "Location: [direction]" where direction must be one of: "North", "Northeast", "East", "Southeast", "South", "Southwest", "West", "Northwest", "Center" (e.g., "Location: Northeast")\\
   3. If there are units visible, prioritize using unit as region of interest.\\
   4. If the target\_id is required, you MUST only use enemy/ally's unit\_ids that are currently visible in your shooting range.\\
   5. If your chosen region of interest is location, you MUST verify its availability.\\
   6. If shared minimap information reveals enemies outside your sight range, prioritize moving to those locations unless there are enemies within your current vision range.\\
   7. Your chosen region of interest should align with the current task description and ally's intentions.\\
   8. Your chosen region of interest should enable you to quickly engage in combat or efficiently achieve the task in cooperation with allies?\\
Reasoning of region of interest: \\
Why was this region of interest chosen? \\
You should only respond in the format described below with a line break after each section colon (\#\#Section\#\#:) and NOT output comments or other information:\\
\#\#Game\_situation\#\#:\\
1. ...\\
\#\#Region\_of\_interest\#\#:\\
region of interest\\
\#\#Reasoning\_of\_region\_of\_interest\#\#:\\
1. ...
}
\end{tcolorbox}

\begin{tcolorbox}[breakable,colback=black!5,colframe=black!40!black,title=Prompt for Task Reasoning]
\small{
You are an AI assistant helping with academic research in the StarCraft II's SMAC (StarCraft Multi-Agent Challenge) environment, controlling a $<$unit\_type$>$ unit with ID <$unit_id$> in micromanagement scenarios $<$scenario\_name$>$ to help your team defeat the enemy forces. You operate under decentralized execution with partial observability, making decisions based only on local observations within your unit's field of view. You will be sequentially given $<$event\_count$>$ screenshots and corresponding descriptions of recent events. You will also be given a summary of the history that happened before the last screenshot. By analyzing these inputs, you gain a comprehensive understanding of the current context and situation within the game. You should assist in summarizing the next immediate task to do in SMACv2. Your ultimate goal is to help your team defeat the enemy forces as quickly as possible.\\
I will give you the following information:\\
Reasoning for the last episode:\\
$<$last\_episode\_reasoning$>$\\
Cumulative reward for the executing skill:\\
$<$cumulative\_reward$>$\\
Current task:\\
$<$task\_description$>$\\
Ally's tasks:\\
$<$ally\_task$>$\\
Minimap information:\\
$<$ego\_minimap$>$\\
Current game situation:\\
$<$game\_situation$>$\\
Tactics recommendation:\\
$<$web\_search$>$\\
The following are successive screenshots:\\
$<$image\_introduction$>$\\
Skill set in Python format to select the next skill:\\
$<$skill\_library$>$\\
Current executing skill:\\
$<$previous\_action$>$\\
Implementation of the skill:\\
$<$action\_code$>$\\
Reasoning for the skill:\\
$<$previous\_reasoning$>$\\
Self-reflection for the last executed skill:\\
$<$previous\_self\_reflection\_reasoning$>$\\
Task\_guidance: \\
Based on the comprehensive game state analysis and team context, decompose the primary objective of "defeat all enemy units" into ONE specific tactical sub-task that enhances either target prioritization (score\_target) or behavior control (control\_logic). This sub-task should be concrete, implementable, and aligned with team coordination.
Consider the following in your task decomposition:\\
1. Final Objective: Defeat enemy forces while preserving allies\\
2. Team Context:\\
   - Your unit's current assigned task
   - Ally units' assigned tasks
   - Progress made on previous tasks
3. Tactical Layer:\\
   - Enemy unit compositions and strategies
   - Team formation and positioning
The task should follow one of these formats:\\
For target prioritization (score\_target):\\
"Adjust [scoring weight/multiplier/threshold] to [specific combat calculation] based on [unit composition + battle state] where [precise condition]"\\
For behavior control (control\_logic):\\
"Implement [unit movement pattern/formation/targeting] when [combat state + ally positions] satisfy [precise conditions]"
Task Requirements:\\
Specificity: Must define exact behavior modification
Measurability: Must have clear success criteria
Actionability: Must be achievable using available atomic actions
Coordination: Must support team tactical objectives
Adaptability: Must respond to changing battle conditions
If current task implementation remains unsuccessful, output 'null'.\\
Reasoning\_of\_task: \\
Why was this new task chosen, or why is there no need to propose a new task? \\
Skill\_guidance:\\
Based on the current executing skill and the proposed next task, evaluate if there is alignment between them. Output True if the current skill effectively supports the task requirements, or False if a new skill is needed.\\
Reasoning\_of\_skill: \\
Why was this decision chosen? \\
You should only respond in the format described below with a line break after each section colon (\#\#Section\#\#:) and NOT output comments or other information:\\
\#\#Task\_guidance\#\#:

[task guidance]

\#\#Skill\_guidance\#\#:

[True or False]
}
\end{tcolorbox}

\begin{tcolorbox}[breakable,colback=black!5,colframe=black!40!black,title=Prompt for Skill Generation]
\small{
You are an AI assistant helping with academic research in the StarCraft II's SMAC (StarCraft Multi-Agent Challenge) environment, controlling a $<$unit\_type$>$ unit with ID <$unit_id$> in micromanagement scenarios $<$scenario\_name$>$ to help your team defeat the enemy forces. You operate under decentralized execution with partial observability, making decisions based only on local observations within your unit's field of view. Your task is to enhance combat effectiveness:\\
Reasoning for the last episode:\\
$<$last\_episode\_reasoning$>$\\
Cumulative reward for the executing skill:\\
$<$cumulative\_reward$>$\\
Current task:\\
$<$task\_description$>$\\
Ally's tasks:\\
$<$ally\_task$>$\\
Minimap information:\\
$<$ego\_minimap$>$\\
Current game situation:\\
$<$game\_situation$>$\\
$<$image\_introduction$>$\\
Skill set in Python format to select the next skill:\\
$<$skill\_library$>$\\
Current executing skill:\\
$<$previous\_action$>$\\
Implementation of the skill:\\
$<$action\_code$>$\\
Reasoning for the skill:\\
$<$previous\_reasoning$>$\\
Self-reflection for the last executed skill:\\
$<$previous\_self\_reflection\_reasoning$>$\\
Combat Analysis Task:\\
1. Analyze the provided script's effectiveness\\
2. Analyze the score\_target(unit) function's effectiveness and weaknesses.\\
3. Analyze the control\_logic() function's effectiveness and weaknesses.\\
4. Based on the current executing skill, the existing skills in skill library, and current task, evaluate if there is alignment between them. \\
5. If a new skill is needed, design tactical improvements while maintaining code structure.\\
6. If the current skill or there is any skill in skill library effectively supports the task requirements, output 'null' to avoid unnecessary token consumption.\\
Identify critical function for improvement (choose ONE Prioritize score\_target(unit)):\\
1. score\_target(unit): Target priority and scoring system. (Preferred)\\
2. control\_logic(): Unit movement and attack decision making.\\
Skill\_generation: \\
If there is no enemies, only output 'null'.\\
If the current skill or there is any skill in skill library effectively supports the task requirements, only output 'null'. \\
Otherwise:\\
The content of the improved code should obey the following code rules:\\
    1. Output Format: Only provide the complete improved function (score\_target(unit) (Preferred) OR control\_logic()).\\
    2. If the improved function is score\_target(unit), there is exactly one parameter named "unit".\\
    3. If the improved function is control\_logic(), it should take no parameters.\\
    4. The code should be surrounded in the '```python' and '```' structure. \\

You should only respond in the format described below with a line break after each section colon (\#\#Section\#\#:) and NOT output comments or other information:\\

\#\#Skill\_generation\#\#:\\
```python

def [function\_name]([parameters]):

    [improved implementation]
```
}
\end{tcolorbox}

\begin{tcolorbox}[breakable,colback=black!5,colframe=black!40!black,title=Prompt for Actor]
\small{
You are an AI assistant helping with academic research in the StarCraft II's SMAC (StarCraft Multi-Agent Challenge) environment, controlling a $<$unit\_type$>$ unit with ID <$unit_id$> in micromanagement scenarios $<$scenario\_name$>$ to help your team defeat the enemy forces. You operate under decentralized execution with partial observability, making decisions based only on local observations within your unit's field of view. Utilizing this insight, you are tasked with identifying the most suitable skill to take next, given the current task. You control the game unit and can execute skills from the available skill set. Upon evaluating the provided information, your role is to articulate the precise skill you would deploy, considering the game's present circumstances, and specify any necessary parameters for implementing that skill:\\
$<$last\_episode\_reasoning$>$\\
Cumulative reward for the executing skill:\\
$<$cumulative\_reward$>$\\
Current task:\\
$<$task\_description$>$\\
Ally's tasks:\\
$<$ally\_task$>$\\
Minimap information:\\
$<$ego\_minimap$>$\\
Current game situation:\\
$<$game\_situation$>$\\
$<$image\_introduction$>$\\
Skill set in Python format to select the next skill:\\
$<$skill\_library$>$\\
Current executing skill:\\
$<$previous\_action$>$\\
Implementation of the skill:\\
$<$action\_code$>$\\
Reasoning for the skill:\\
$<$previous\_reasoning$>$\\
Self-reflection for the last executed skill:\\
$<$previous\_self\_reflection\_reasoning$>$\\
Skills: \\
The best skill to execute next to progress in achieving the goal. Pay attention to the names of the available skills and to the previous skills already executed, if any. You should also pay more attention to the following skill rules:\\
    1. ONLY choose skill in the provided skill set.\\
    2. Output skills in Python code format with required keyword parameters.\\
    3. The ONLY required keyword parameter is "obs: str" - you MUST include this parameter as "obs='current'" in every skill. The actual observation will be automatically injected at runtime.\\
    4. If there is summarization of history, consider this information when selecting the skill.\\
    5. If the error report indicates that the last skill was unavailable, you MUST select a different skill.\\
    6. Consider coordination with other units and choose skills that enhance team performance and cooperation.\\
    7. Avoid repeating the same skill as the last executed skill unless there is a compelling strategic reason.\\
You should only respond in the format described below with a line break after each section colon (\#\#Section\#\#:) and NOT output comments or other information:\\

\#\#Skills\#\#:\\
```python

    skill\_name(obs='current')
    
```
}
\end{tcolorbox}
\begin{lstlisting}[
    style=mystyle,
    caption={Example Script of Skill Initialzation},
    label=code:long
]
def race_melee_ranged_medivac_navi_A_star_score_type_default_center(obs: str):
    """
    Zealot/Zergling/Baneling/Colossus/Stalker/Hydralisk/Marauder/Marine/Medivac Controls Logic:
        Medivac:
            - Heals allies below 100% HP
            - Maintains 0.75 sight range from enemies
            - Centers between allies when no healing targets

        Melee (Zealot/Zergling/Baneling):
            - Attacks highest threat target within 0.7 sight range
            - Pursues targets using A* pathfinding
            - Groups with allies at >0.7 distance threshold

        Ranged (Colossus/Stalker/Hydralisk/Marauder/Marine):
            - Kites melee units at max_range - 0.05 
            - Focus fires targets shared with 2+ allies
            - Maintains position behind melee allies

        Key Implementation:
        - Pathfinding: A* pathfinding in 32x32 grid with unit collision radius
        - Target scoring: [0-10] based on type (colossus 9 > baneling 8 > zealot 7 > stalker 6 > hydralisk 5 > marauder 4 > marine 3 > zergling 2 > medivac 1)/health (0-0.6)/distance (0-0.3)/last attacked (0.1)
        - Default action: Move to center of map, parse region of interest, random choice

    Args:
        obs (str): Observation string containing game state
    """
    import math
    # Parse observation
    obs_data = parse_obs(obs)
    # Get set of available actions
    valid_actions = obs_data.available_actions

    if 0 in valid_actions:
        return 0    
    
    def score_target(unit):
        """Enhanced target scoring with improved kiting and formation control"""
        if unit.health <= 0:
            return -1
                    
        score = 0
        
        # Refined unit type priorities with enhanced threat scaling
        unit_priorities = {
            'colossus': 35.0,  # Further increased priority
            'stalker': 30.0,   # Enhanced anti-armor focus
            'zealot': 45.0,    # Higher melee threat recognition

            'marine': 45.0,    # Balanced damage dealer priority
            'marauder': 35.0,  # Anti-armor specialist
            'medivac': 30.0,   # Support unit priority

            'hydralisk': 30.0,  # High priority for their sustained DPS
            'zergling': 35.0,   # Medium priority as swarm units
            'baneling': 45.0,   # Critical priority due to splash damage
        }
        
        # Dynamic matchup priorities with improved counter weighting
        unit_counters = {
            'colossus': {'colossus': 1.2, 'stalker': 1.0, 'zealot': 1.5},
            'stalker': {'colossus': 1.2, 'stalker': 1.0, 'zealot': 1.5},
            'zealot': {'colossus': 1.2, 'stalker': 1.0, 'zealot': 1.5},

            'marine': {'marine': 1.5, 'medivac': 1.0, 'marauder': 1.2},
            'marauder': {'marine': 1.5, 'medivac': 1.0, 'marauder': 1.2},
            'medivac': {'marine': 1.5, 'medivac': 1.0, 'marauder': 1.2},

            'hydralisk': {'hydralisk': 1.0, 'zergling': 1.2, 'baneling': 1.5},
            'zergling': {'hydralisk': 1.2, 'zergling': 1.5, 'baneling': 1.0},
            'baneling': {'hydralisk': 1.2, 'zergling': 1.5, 'baneling': 1.0},

        }
        
        base_priority = unit_priorities.get(unit.unit_type.lower(), 5.0)
        is_ranged = unit.unit_type.lower() not in ['zealot', 'zergling', 'baneling']
        own_is_ranged = obs_data.own_unit_type.lower() not in ['zealot', 'zergling', 'baneling']
        
        # Enhanced threat assessment with improved melee handling
        matchup_mult = unit_counters.get(obs_data.own_unit_type.lower(), {}).get(unit.unit_type.lower(), 1.0)
        base_priority *= matchup_mult
        
        if hasattr(unit, 'can_attack'):  # Enemy unit
            score = base_priority

            distance_factor = max((1 - unit.distance) + 1, 0.5)
            score *= distance_factor

            if not own_is_ranged:
                range_ally = [ally for ally in obs_data.allies if ally.unit_type.lower() not in ['zealot', 'zergling', 'baneling']]
                if range_ally:
                    ally_x = sum(ally.position[0] for ally in range_ally) / len(range_ally)
                    ally_y = sum(ally.position[1] for ally in range_ally) / len(range_ally)
                    ally_distance = ((ally_x - unit.position[0])**2 + (ally_y - unit.position[1])**2)**0.5
                    distance_factor = max((1 - ally_distance) + 1, 0.5)
                    score *= distance_factor
            
            # Enhanced Position Analysis with improved spacing
            position_x, position_y = unit.position
            
            def calculate_combat_power(units, radius=0.5):  # Further reduced for tighter control
                total_power = 0
                ranged_count = 0
                melee_count = 0
                unit_positions = []
                
                for u in units:
                    dist = ((u.position[0] - position_x)**2 + 
                        (u.position[1] - position_y)**2)**0.5
                    unit_positions.append(u.position)
                    
                    if dist <= radius:
                        base_power = unit_priorities.get(u.unit_type.lower(), 5.0)
                        
                        # Unit type specific power calculation
                        if u.unit_type.lower() in ['zealot', 'zergling', 'baneling']:
                            melee_count += 1
                            if melee_count >= 2:
                                base_power *= 1.4
                        else:
                            ranged_count += 1
                            base_power *= 1.3
                        
                        # Health-based power scaling
                        health_factor = 1.5 if u.health > 0.7 else 1.0 if u.health > 0.4 else 0.6
                        position_factor = 1.3 - (dist/radius)
                        
                        total_power += base_power * health_factor * position_factor
                
                # Enhanced formation cohesion calculation
                cohesion = 0
                if len(unit_positions) > 2:
                    center_x = sum(p[0] for p in unit_positions) / len(unit_positions)
                    center_y = sum(p[1] for p in unit_positions) / len(unit_positions)
                    avg_dist = sum(((p[0] - center_x)**2 + (p[1] - center_y)**2)**0.5 
                                for p in unit_positions) / len(unit_positions)
                    max_desired_dist = 0.3  # Tighter formation control
                    cohesion = 2.0 / (1.0 + (avg_dist / max_desired_dist))
                
                return total_power * (1 + cohesion), melee_count, ranged_count
            
            ally_power, ally_swarms, ally_ranged = calculate_combat_power(obs_data.allies)
            enemy_power, enemy_swarms, enemy_ranged = calculate_combat_power(obs_data.enemies)
            
            # Improved Focus Fire Logic with enhanced commitment
            num_attackers = sum(1 for ally in obs_data.allies 
                            if ally.last_action >= 6 and ally.last_action - 6 == unit.id)
            
            if unit.id == obs_data.last_action - 6:
                persistence_bonus = 2.0  # Stronger target commitment
                score *= persistence_bonus
            
            if num_attackers > 0:
                focus_bonus = 1.2 ** num_attackers  # Enhanced focus fire emphasis
                # if num_attackers too high, discourage prevent overcommitment
                if num_attackers >= 3 and unit.id != obs_data.last_action - 6 and obs_data.own_unit_type.lower() in ['zergling', 'baneling']:
                    focus_bonus = 0.5
                score *= focus_bonus
            
            # Improved Combat Advantage Factor
            advantage_factor = 1.0
            if ally_power > enemy_power * 1.3:
                advantage_factor = 1.2  # More aggressive advantage pursuit
                if ally_swarms >= 3:
                    advantage_factor *= 1.2
            
            # prioritize isolated enemies
            if (enemy_swarms + enemy_ranged) == 1:
                advantage_factor *= 2.0
            elif (ally_swarms + ally_ranged) > (enemy_swarms + enemy_ranged):
                advantage_factor *= 1.2

            score *= advantage_factor
            
            # Health factor
            health_factor = (1 - unit.health) + 1
            score *= health_factor
            
        else:  # Ally unit
            score = base_priority
            
            # Improved Support Priority
            health_factor = (1 - unit.health) + 1
            score *= health_factor
            
            distance_factor = max((1 - unit.distance) + 1, 0.5)
            score *= distance_factor
            
        return score

    def control_logic():
        # Medivac units control logic
        if obs_data.own_unit_type.lower() == 'medivac':
            attack_actions = [a for a in valid_actions if a >= 6]
            # If there are allies
            if obs_data.allies:
                lowest_health_ally = min(obs_data.allies, key=lambda x: x.health)
                # If there are both allies and enemies
                if obs_data.enemies:
                    enemy_in_range = [enemy for enemy in obs_data.enemies if enemy.distance < 1]
                    # Check if any melee ally, if so and last action is not attack, move to center of melee allies
                    melee_ally = [ally for ally in obs_data.allies if ally.unit_type.lower() in ['zealot', 'zergling', 'baneling']]
                    if melee_ally:
                        # Move to center of melee allies
                        ally_x = sum(ally.position[0] for ally in melee_ally) / len(melee_ally)
                        ally_y = sum(ally.position[1] for ally in melee_ally) / len(melee_ally)
                    else:
                        ally_x = sum(ally.position[0] for ally in obs_data.allies) / len(obs_data.allies)
                        ally_y = sum(ally.position[1] for ally in obs_data.allies) / len(obs_data.allies)
                    # Calculate retreat position
                    if len(enemy_in_range) > 0:
                        enemy_center = (sum(e.position[0] for e in enemy_in_range) / len(enemy_in_range),
                                    sum(e.position[1] for e in enemy_in_range) / len(enemy_in_range))
                    else:
                        enemy_center = (sum(e.position[0] for e in obs_data.enemies) / len(obs_data.enemies),
                                    sum(e.position[1] for e in obs_data.enemies) / len(obs_data.enemies))
                    
                    dx = ally_x - enemy_center[0]
                    dy = ally_y - enemy_center[1]

                    distance = (dx ** 2 + dy ** 2) ** 0.5

                    safe_x = ally_x + (dx / abs(dx)) * 2/obs_data.own_sight_range if dx != 0 else ally_x
                    safe_y = ally_y + (dy / abs(dy)) * 2/obs_data.own_sight_range if dy != 0 else ally_y

                    distance = (safe_x ** 2 + safe_y ** 2) ** 0.5
                    criterion = 5/obs_data.own_sight_range
                    if obs_data.last_action >= 6 or len(attack_actions) == 0:
                        target_angle = math.atan2(enemy_center[1], enemy_center[0])
                        safe_angle = math.atan2(ally_y, ally_x)
                        angle_diff = abs(target_angle - safe_angle)
                        if distance > criterion and (math.pi/9 < angle_diff < 17*math.pi/9):
                            path_action = find_path(obs_data, safe_x, safe_y)
                            if path_action:
                                return path_action
                target_scores = {ally.id: score_target(ally) for ally in obs_data.allies}
                # Check if there are same max score targets
                max_score = max(target_scores.values())
                max_score_target_ids = [target_id for target_id, score in target_scores.items() if score == max_score]
                max_score_targets = [ally for ally in obs_data.allies if ally.id in max_score_target_ids]
                closest_ally = min(obs_data.allies, key=lambda x: x.distance)
                if len(max_score_targets) > 1:
                    # Chose the closest target
                    best_target = min(max_score_targets, key=lambda x: x.distance)
                else:
                    best_target = max_score_targets[0]
                if (best_target.id + 6) in valid_actions and 0 < best_target.health < 0.9:
                    return heal(best_target.id)
                elif (closest_ally.id + 6) in valid_actions and 0 < closest_ally.health < 0.9:
                    return heal(closest_ally.id)
                elif (lowest_health_ally.id + 6) in valid_actions and 0 < lowest_health_ally.health < 0.9:
                    return heal(lowest_health_ally.id)
                else:
                    # Move to the target
                    dx = best_target.position[0]
                    dy = best_target.position[1]
                    path_action = find_path(obs_data, dx, dy, target_type=best_target.unit_type.lower())
                    if path_action:
                        return path_action
            # If there are no allies
            else:
                # If there are only enemies
                if obs_data.enemies:
                    enemy_x = sum(e.position[0] for e in obs_data.enemies) / len(obs_data.enemies)
                    enemy_y = sum(e.position[1] for e in obs_data.enemies) / len(obs_data.enemies)
                    target_x = - enemy_x
                    target_y = - enemy_y

                    g_x = target_x * obs_data.own_sight_range + (obs_data.own_position[0] * 32)
                    g_y = target_y * obs_data.own_sight_range + (obs_data.own_position[1] * 32)
                    if not (0 <= g_x <= 32 and 0 <= g_y <= 32):
                        target_x = (0.5 - obs_data.own_position[0]) * 32 / obs_data.own_sight_range
                        target_y = (0.5 - obs_data.own_position[1]) * 32 / obs_data.own_sight_range

                    path_action = find_path(obs_data, target_x, target_y)
                    if path_action:
                        return path_action
        # Melee units control logic
        elif obs_data.own_unit_type.lower() in ['zealot', 'zergling', 'baneling']:
            # If there are enemies
            if obs_data.enemies:
                enemy_in_range = [enemy for enemy in obs_data.enemies if enemy.distance < 1]
                attack_actions = [a for a in valid_actions if a >= 6]

                if len(enemy_in_range) > 0:
                    enemy_center = (sum(e.position[0] for e in enemy_in_range) / len(enemy_in_range),
                                sum(e.position[1] for e in enemy_in_range) / len(enemy_in_range))
                else:
                    enemy_center = (sum(e.position[0] for e in obs_data.enemies) / len(obs_data.enemies),
                                sum(e.position[1] for e in obs_data.enemies) / len(obs_data.enemies))
                if obs_data.allies:
                    # Check if any melee ally, if so and last action is not attack, move to center of melee allies
                    melee_ally = [ally for ally in obs_data.allies if ally.unit_type.lower() in ['zealot', 'zergling', 'baneling']]
                    if melee_ally:
                        # Move to center of melee allies
                        ally_x = sum(ally.position[0] for ally in melee_ally) / len(melee_ally) / 2
                        ally_y = sum(ally.position[1] for ally in melee_ally) / len(melee_ally) / 2

                        safe_x = ally_x 
                        safe_y = ally_y

                        distance = (safe_x ** 2 + safe_y ** 2) ** 0.5
                        criterion = 2/obs_data.own_sight_range
                        if len(attack_actions) == 0 or distance > 0.5:
                            target_angle = math.atan2(enemy_center[1], enemy_center[0])
                            safe_angle = math.atan2(ally_y, ally_x)
                            angle_diff = abs(target_angle - safe_angle)
                            if distance > criterion and (math.pi/9 < angle_diff < 17*math.pi/9 or distance > 0.5):
                                path_action = find_path(obs_data, safe_x, safe_y)
                                if path_action:
                                    return path_action
                            
                # Enhanced cluster detection with dynamic radius
                enemy_clusters = {}
                cluster_centers = {}
                for enemy in obs_data.enemies:
                    nearby_enemies = []
                    center_x, center_y = enemy.position[0], enemy.position[1]
                    
                    # Dynamic cluster radius based on unit type
                    cluster_radius = 0.3 if obs_data.own_unit_type.lower() == 'baneling' else 0.2
                    
                    for other in obs_data.enemies:
                        distance = ((other.position[0] - enemy.position[0])**2 + 
                                (other.position[1] - enemy.position[1])**2)**0.5
                        if distance <= cluster_radius:
                            nearby_enemies.append(other)
                            center_x += other.position[0]
                            center_y += other.position[1]
                    
                    if nearby_enemies:
                        center_x /= len(nearby_enemies)
                        center_y /= len(nearby_enemies)
                        
                    enemy_clusters[enemy.id] = len(nearby_enemies)
                    cluster_centers[enemy.id] = (center_x, center_y)

                # Enhanced target scoring with tactical considerations
                target_scores = {}
                for enemy in obs_data.enemies:
                    base_score = score_target(enemy)
                    
                    # Enhanced cluster bonus for splash damage
                    if obs_data.own_unit_type.lower() == 'baneling':
                        cluster_bonus = 1.5 ** enemy_clusters[enemy.id]
                    else:
                        cluster_bonus = 1.2 ** enemy_clusters[enemy.id]
                    # Calculate final score with all factors
                    target_scores[enemy.id] = (base_score + cluster_bonus)

                # Check if there are same max score targets
                max_score = max(target_scores.values())
                max_score_targets = [enemy for enemy in obs_data.enemies 
                           if target_scores[enemy.id] >= max_score]  # Allow for close scores
                if len(max_score_targets) > 1:
                    # Choose target balancing distance and cluster potential
                    best_target = min(max_score_targets, 
                            key=lambda x: x.distance)
                else:
                    best_target = max_score_targets[0]

                if best_target.can_attack:
                    return attack(best_target.id)
                else:
                    # Move to the target
                    dx = best_target.position[0]
                    dy = best_target.position[1]
                    path_action = find_path(obs_data, dx, dy, target_type=best_target.unit_type.lower())
                    if path_action:
                        return path_action
                    elif attack_actions:
                        attackable_enemies = [enemy for enemy in obs_data.enemies if enemy.can_attack]
                        if obs_data.last_action in attack_actions:
                            return obs_data.last_action
                        if attackable_enemies:
                            return attack(min(attackable_enemies, key=lambda e: e.distance).id)
                        return random.choice(attack_actions)

            # If there are no enemies
            else:
                # If there are only allies
                if obs_data.allies:
                    # Improved melee group formation
                    melee_allies = [ally for ally in obs_data.allies 
                                if ally.unit_type.lower() in ['zealot', 'zergling', 'baneling']]
                    if melee_allies:
                        spacing = 0.1 if obs_data.own_unit_type.lower() == 'baneling' else 0.05
                        # Dynamic group positioning
                        center_x = sum(ally.position[0] for ally in melee_allies) / len(melee_allies)
                        center_y = sum(ally.position[1] for ally in melee_allies) / len(melee_allies)
                        
                        # Calculate spread from center
                        max_spread = max(((ally.position[0] - center_x)**2 + 
                                        (ally.position[1] - center_y)**2)**0.5 
                                    for ally in melee_allies)
                        
                        own_distance = ((center_x)**2 + (center_y)**2)**0.5
                        
                        if own_distance > spacing or max_spread > 0.1:
                            # Move toward center while maintaining minimum spacing
                            adjusted_x = center_x * 0.85  # Slight offset to prevent overcrowding
                            adjusted_y = center_y * 0.85
                            path_action = find_path(obs_data, adjusted_x, adjusted_y)
                            if path_action:
                                return path_action
                    else:
                        ally_x = sum(ally.position[0] for ally in obs_data.allies) / len(obs_data.allies)
                        ally_y = sum(ally.position[1] for ally in obs_data.allies) / len(obs_data.allies)
                        distance = (ally_x ** 2 + ally_y ** 2) ** 0.5
                        if distance > 0.05:
                            dx = ally_x
                            dy = ally_y
                            path_action = find_path(obs_data, dx, dy)
                            if path_action:
                                return path_action
        # Ranged units control logic
        else:
            attack_actions = [a for a in valid_actions if a >= 6]
            # If there are enemies
            if obs_data.enemies:
                # If there are both allies and enemies
                # Calculate retreat position
                enemy_in_range = [enemy for enemy in obs_data.enemies if enemy.distance < 1]
                if len(enemy_in_range) > 0:
                    enemy_center = (sum(e.position[0] for e in enemy_in_range) / len(enemy_in_range),
                                sum(e.position[1] for e in enemy_in_range) / len(enemy_in_range))
                else:
                    enemy_center = (sum(e.position[0] for e in obs_data.enemies) / len(obs_data.enemies),
                                sum(e.position[1] for e in obs_data.enemies) / len(obs_data.enemies))
                if obs_data.allies:
                    # Check if any melee ally, if so and last action is not attack, move to center of melee allies
                    melee_ally = [ally for ally in obs_data.allies if ally.unit_type.lower() in ['zealot', 'zergling', 'baneling']]
                    melee_enemy = [enemy for enemy in enemy_in_range if enemy.unit_type.lower() in ['zealot', 'zergling', 'baneling']]
                    if melee_ally:
                        # Move to center of melee allies
                        ally_x = sum(ally.position[0] for ally in melee_ally) / len(melee_ally)
                        ally_y = sum(ally.position[1] for ally in melee_ally) / len(melee_ally)
                        
                    else:
                        ally_x = sum(ally.position[0] for ally in obs_data.allies) / len(obs_data.allies) / 2
                        ally_y = sum(ally.position[1] for ally in obs_data.allies) / len(obs_data.allies) / 2

                    dx = ally_x - enemy_center[0]
                    dy = ally_y - enemy_center[1]
                    safe_x = ally_x
                    safe_y = ally_y
                    if melee_ally:
                        safe_x = safe_x + (dx / abs(dx)) * 2/obs_data.own_sight_range if dx != 0 else safe_x
                        safe_y = safe_y + (dy / abs(dy)) * 2/obs_data.own_sight_range if dy != 0 else safe_y
                    melee_threaten = False
                    if melee_enemy:
                        closest_melee_enemy = min(melee_enemy, key=lambda x: x.distance)
                        if closest_melee_enemy.distance <= 4/obs_data.own_sight_range:
                            melee_threaten = True
                            dx = safe_x - closest_melee_enemy.position[0]
                            dy = safe_y - closest_melee_enemy.position[1]
                            safe_x = safe_x + (dx / abs(dx)) * 1/obs_data.own_sight_range if dx != 0 else safe_x
                            safe_y = safe_y + (dy / abs(dy)) * 1/obs_data.own_sight_range if dy != 0 else safe_y
                    
                    distance = (safe_x ** 2 + safe_y ** 2) ** 0.5
                    criterion = 4/obs_data.own_sight_range
                    if obs_data.last_action >= 6 or len(attack_actions) == 0 or distance > 0.9:
                        target_angle = math.atan2(enemy_center[1], enemy_center[0])
                        safe_angle = math.atan2(ally_y, ally_x)
                        angle_diff = abs(target_angle - safe_angle)
                        if distance > criterion and ((math.pi/9 < angle_diff < 17*math.pi/9) or melee_threaten or distance > 0.9):
                            path_action = find_path(obs_data, safe_x, safe_y)
                            if path_action:
                                return path_action
                    # Focus fire logic
                    # Count how many allies are attacking each enemy
                    target_counts = {}
                    for ally in obs_data.allies:
                        if ally.last_action >= 6:
                            target_id = ally.last_action - 6
                            target_counts[target_id] = target_counts.get(target_id, 0) + 1
                    # Distance to safe point of each enemy affect target choosing
                    enemy_safe_distance = {enemy.id: ((enemy.position[0] - safe_x) ** 2 + (enemy.position[1] - safe_y) ** 2) ** 0.5 for enemy in obs_data.enemies}
                    # Find best target combining focus fire and threat scoring
                    target_scores = {enemy.id: score_target(enemy) for enemy in obs_data.enemies}
                    for target_id, count in target_counts.items():
                        if target_id in target_scores:
                            target_scores[target_id] += count * 0.5
                    for target_id, scores in target_scores.items():
                        target_scores[target_id] = scores * (1 - enemy_safe_distance[target_id] * 0.3)
                    best_target_id = max(target_scores.items(), key=lambda x: x[1])[0]
                    best_target = next(enemy for enemy in obs_data.enemies if enemy.id == best_target_id)
                    if best_target.can_attack:
                        return attack(best_target_id)
                    else:
                        # Best target is not in shoot range, move to target
                        dx = best_target.position[0]
                        dy = best_target.position[1]

                        # Only move to target if its direction is not conflicting with the safe point
                        # Check if target direction aligns with safe point direction
                        target_angle = math.atan2(dy, dx)
                        safe_angle = math.atan2(ally_y, ally_x)
                        angle_diff = abs(target_angle - safe_angle)
                        # Only move if angle difference is less than 90 degrees
                        if angle_diff < math.pi/9 or angle_diff > 17*math.pi/9 or not melee_ally:
                            if best_target.distance > obs_data.own_shoot_range / obs_data.own_sight_range:
                                path_action = find_path(obs_data, dx, dy, target_type=best_target.unit_type.lower())
                                if path_action:
                                    return path_action
                        if attack_actions:
                            if obs_data.last_action in attack_actions:
                                return obs_data.last_action
                            attackable_enemies = [enemy for enemy in obs_data.enemies if enemy.can_attack]
                            closest_enemy = min(
                                [enemy for enemy in attackable_enemies],
                                key=lambda enemy: enemy.distance,
                                default=None,
                            )
                            if closest_enemy and closest_enemy.can_attack:
                                return attack(closest_enemy.id)
                            return random.choice(attack_actions)
                        else:
                            if distance > criterion:
                                path_action = find_path(obs_data, safe_x, safe_y)
                                if path_action:
                                    return path_action
                        
                # If there are only enemies
                else:
                    # Closest enemy as target
                    closest_enemy = min(
                        [enemy for enemy in obs_data.enemies],
                        key=lambda enemy: enemy.distance,
                        default=None,
                    )
                    # No allies, kitting melee enemies
                    if closest_enemy.unit_type.lower() in ['zealot', 'zergling', 'baneling']:
                        if closest_enemy.distance <= 4 / obs_data.own_sight_range and obs_data.last_action >= 6:
                            enemy_x = sum(e.position[0] for e in obs_data.enemies) / len(obs_data.enemies)
                            enemy_y = sum(e.position[1] for e in obs_data.enemies) / len(obs_data.enemies)
                            target_x = - enemy_x
                            target_y = - enemy_y

                            g_x = target_x * obs_data.own_sight_range + (obs_data.own_position[0] * 32)
                            g_y = target_y * obs_data.own_sight_range + (obs_data.own_position[1] * 32)
                            if not (0 <= g_x <= 32 and 0 <= g_y <= 32):
                                target_x = (0.5 - obs_data.own_position[0]) * 32 / obs_data.own_sight_range
                                target_y = (0.5 - obs_data.own_position[1]) * 32 / obs_data.own_sight_range

                            path_action = find_path(obs_data, target_x, target_y)
                            if path_action:
                                return path_action
                        if closest_enemy.can_attack:
                            return attack(closest_enemy.id)
                    else:
                        # No melee enemies, highest priority enemy as target
                        target_scores = {enemy.id: score_target(enemy) for enemy in obs_data.enemies}
                        # Check if there are same max score targets
                        max_score = max(target_scores.values())
                        max_score_target_ids = [target_id for target_id, score in target_scores.items() if score == max_score]
                        max_score_targets = [enemy for enemy in obs_data.enemies if enemy.id in max_score_target_ids]
                        if len(max_score_targets) > 1:
                            # Chose the closest target
                            best_target = min(max_score_targets, key=lambda x: x.distance)
                        else:
                            best_target = max_score_targets[0]
                        if best_target.can_attack:
                            return attack(best_target.id)
                        else:
                            # Best target is not in shoot range, move to target
                            dx = best_target.position[0]
                            dy = best_target.position[1]
                            if best_target.distance > obs_data.own_shoot_range / obs_data.own_sight_range:
                                path_action = find_path(obs_data, dx, dy, target_type=best_target.unit_type.lower())
                                if path_action:
                                    return path_action
                                elif attack_actions:
                                    if obs_data.last_action in attack_actions:
                                        return obs_data.last_action
                                    attackable_enemies = [enemy for enemy in obs_data.enemies if enemy.can_attack]
                                    closest_enemy = min(
                                        [enemy for enemy in attackable_enemies],
                                        key=lambda enemy: enemy.distance,
                                        default=None,
                                    )
                                    if closest_enemy and closest_enemy.can_attack:
                                        return attack(closest_enemy.id)
                                    return random.choice(attack_actions)
            # If there are no enemies
            else:
                # If there are only allies
                if obs_data.allies:
                    # Check if any melee ally
                    melee_ally = [ally for ally in obs_data.allies if ally.unit_type.lower() in ['zealot', 'zergling', 'baneling']]
                    if melee_ally:
                        # Move to center of melee allies
                        melee_ally_x = sum(ally.position[0] for ally in melee_ally) / len(melee_ally)
                        melee_ally_y = sum(ally.position[1] for ally in melee_ally) / len(melee_ally)
                        dx = melee_ally_x
                        dy = melee_ally_y
                        distance = (dx ** 2 + dy ** 2) ** 0.5
                        if distance > 0.05:
                            path_action = find_path(obs_data, dx, dy)
                            if path_action:
                                return path_action
                    else:
                        # No melee allies, move to target ally
                        ally_x = sum(ally.position[0] for ally in obs_data.allies) / len(obs_data.allies)
                        ally_y = sum(ally.position[1] for ally in obs_data.allies) / len(obs_data.allies)
                        distance = (ally_x ** 2 + ally_y ** 2) ** 0.5
                        if distance > 0.05:
                            dx = ally_x
                            dy = ally_y
                            path_action = find_path(obs_data, dx, dy)
                            if path_action:
                                return path_action
        return default_action(obs)

    return control_logic()
\end{lstlisting}

\end{document}